\documentclass[final,5p,times,twocolumn]{elsarticle}

\usepackage[T1]{fontenc}
\usepackage[utf8]{inputenc}
\usepackage{amsmath,amssymb,amsfonts}
\usepackage{booktabs}
\usepackage{graphicx}
\usepackage{enumitem}
\usepackage{xcolor}
\usepackage{hyperref}
\usepackage{multirow}
\usepackage{array}
\usepackage{subcaption}
\usepackage{tikz}
\usepackage{makecell}
\usepackage{dblfloatfix}
\usepackage{algorithm}
\usepackage{algpseudocode}
\usetikzlibrary{positioning,arrows.meta,calc,fit,backgrounds,shapes.geometric}
\usepackage{placeins} 
\biboptions{numbers,sort&compress}

\hypersetup{colorlinks=true,linkcolor=blue,citecolor=blue,urlcolor=blue}

\begin{document}

\begin{frontmatter}

\title{Explainable Wastewater Digital Twins:\\
Adaptive Context-Conditioned Structured Simulators with
Self-Falsifying Decision Support}

\author[aau]{Gary Simethy\corref{cor1}}
\ead{gasi@energy.aau.dk}
\author[aau]{Daniel Ortiz Arroyo}
\author[aau]{Petar Durdevic}

\cortext[cor1]{Corresponding author}
\affiliation[aau]{organization={Department of Energy, Aalborg University},
  addressline={Niels Bohrs Vej 8},
  city={Esbjerg},
  postcode={6700},
  country={Denmark}}

\begin{abstract}
Operators of safety-critical industrial processes increasingly rely on
digital twins to screen control interventions, but such simulators
rarely carry certified safety guarantees.  Wastewater treatment plants
exemplify the gap: operators face a daily safety--efficiency trade-off
where aerating too little risks effluent violations and nitrous-oxide
(N\textsubscript{2}O) spikes, and aerating too much wastes energy.
We develop an explainable digital twin for aeration and dosing setpoints.
CCSS-IX, the simulator, is a bank of interpretable locally linear
state-space ``experts'' adaptively mixed by a context-aware gating
network, building on a continuous-time regime-switching
scaffold~\citep{simethy2026ccssrs}.  A runtime decision layer applies
conformal risk control to abstain, reopen, or return a falsifying
temporal witness for any operator-proposed action that cannot be
statistically certified.  The artificial-intelligence
contribution is twofold: an identifiable, context-conditioned structured
surrogate that retains operator-readable dynamics, and a self-falsifying
decision rule with finite-sample coverage guarantees.  The engineering
contribution is a validated, end-to-end decision-support pipeline,
tested on a 1000-step slice of the Aved{\o}re full-scale plant (42.6\%
sensor missingness, 2-minute sampling), the Agtrup/BlueKolding
full-scale plant in Denmark, and the Benchmark Simulation Model No.~2
(BSM2) international benchmark, under a matched ten-seed protocol.  The static structured ensemble lies within
0.78\% root-mean-square error of an unconstrained black-box reference,
and the adaptive variant within 1.08\%.  The calibrated reopen rule
cuts aggregate two-plant regret by 43.6\% at an unsafe-action cost
weight of 4 and eliminates unsafe chosen actions on the BSM2 main
slice.  Event-aligned temporal witnesses prevent 93 of 187 false-safe
N\textsubscript{2}O approvals, about $4.65\times$ the dyadic baseline
(paired McNemar $p < 10^{-21}$).
\end{abstract}


\end{frontmatter}

\section{Introduction}
\label{sec:intro}

Wastewater treatment plants (WWTPs) consume an estimated 1--3\% of national
electricity in industrialised countries, are accountable for a non-trivial
share of national greenhouse-gas inventories through aeration energy and
process emissions of nitrous oxide (N\textsubscript{2}O), and operate under
tightening effluent discharge regulations.  In this setting, operators
increasingly rely on \emph{digital twins} (learned simulators of plant
dynamics) to screen proposed control interventions before deployment: a
process engineer queries the twin with a candidate aeration profile, dosing
strategy, or set-point change and asks whether the predicted plant response
meets safety and quality objectives~\citep{wang2024dt,rasheed2020dt,tao2019dt,
mehta2025watersector}.  The evaluation task has therefore shifted from
fitting observed plant histories to certifying proposed futures.

This shift introduces two practical challenges that current approaches do not
jointly solve.

\noindent\textbf{Interpretability challenge.}
High-fidelity learned simulators for WWTP processes are typically black-box
recurrent or state-space models~\citep{chen2018neuralode,kidger2020ncde,
hansen2024lstmn2o,mdpi2025symmetryn2o}.  They produce trajectory predictions
and uncertainty estimates, but they do not expose the mechanistic channels
through which proposed controls drive predicted responses.  An operator
receiving a simulator approval cannot inspect how planned actuation, system
state, and exogenous disturbances interact to produce that prediction.  In
high-stakes WWTP operation, where a wrong approved intervention can cause
ammonium effluent violations, energy waste, or N\textsubscript{2}O spikes
with $\approx$265$\times$ CO\textsubscript{2} warming
potential~\citep{ravishankara2009n2o,kampschreur2009n2o}, an unexplainable
approval is difficult to act on responsibly~\citep{rudin2019stop}.  Recent
standardisation activity, notably ISO/IEC TR 5469:2024 on functional safety
of AI systems~\citep{iso5469}, makes this expectation increasingly
explicit.

\noindent\textbf{Certification challenge.}
Predictive accuracy on observed histories does not determine whether a
proposed intervention simulation should be trusted.  A simulator can achieve
good average forecast error while systematically approving unsafe interventions
whenever proposed control paths drift from observed operation.  The standard
mitigation, rejecting interventions outside the historical support envelope,
is necessary but insufficient.  Some unsafe interventions remain supported
because they occur near the observed distribution but produce internally
unstable predictions; conversely, some beneficial novel interventions are
discarded because support-novelty alone is treated as a reason to abstain
\citep{geifman2017selective,angelopoulos2022crc}.

\noindent\textbf{Missed connection.}
These two challenges are coupled in a way the existing literature has not made
explicit.  An interpretable simulator that decomposes predictions into
explicit state, control, and disturbance channels is not only more auditable
for human engineers: its structured decomposition is what makes
self-consistency checks operationally meaningful.  When the same intervention
path is evaluated under different temporal decompositions and the decisions
disagree, the disagreement can be attributed to a specific control event or
disturbance transition \emph{only if} the simulator exposes those channels.  A
black-box simulator can detect internal inconsistency but cannot explain which
part of the proposed intervention caused it.  Interpretable structure is
what permits attributing a self-falsification disagreement to a specific
channel; in this paper we treat the two as one design problem.

\noindent\textbf{Contributions.}
This paper develops that connection in the context of wastewater digital twins
and validates it across three industrial WWTP benchmarks.  Our contributions are:

\begin{enumerate}[leftmargin=2em]
  \item \textbf{CCSS-IX: an adaptive context-conditioned structured
    simulator.}  CCSS-IX bridges two extremes: static structured models
    that are interpretable but rigid, and black-box neural simulators
    that are accurate but opaque.  It builds on the continuous-time
    regime-switching CCSS-RS scaffold \citep{simethy2026ccssrs} and
    replaces its opaque regime experts with update modules that
    decompose each latent transition into an explicit state-coupling
    matrix $A_k(\gamma_t)$, control-influence matrix $B_k(\gamma_t)$,
    disturbance-influence matrix $E_k(\gamma_t)$, and a context-modulated
    additive nonlinear response.  Across ten seeds the structured family
    matches the black-box CCSS-RS reference on paired RMSE
    (\S\ref{sec:res-arch}, Table~\ref{tab:archladder}).

  \item \textbf{A self-falsifying validity layer that turns the structured
    simulator into a decision artefact.}  We wrap CCSS-IX in a
    four-outcome decision layer (accept, abstain, calibrated reopen,
    temporal witness).  The temporal witness evaluates the same
    intervention path under an event-aligned partition and reports a
    falsifying disagreement together with the specific control or
    disturbance event responsible.  The event-aligned partition family
    prevents $4.65\times$ as many false-safe N\textsubscript{2}O
    approvals as a dyadic decomposition (93/187 vs 20/187, McNemar
    $p<10^{-21}$ on the same paired pool).

  \item \textbf{Validation on three industrial WWTP benchmarks.}  We evaluate
    on two real full-scale plants (Aved{\o}re~\citep{hansen2024data} and
    Agtrup/BlueKolding~\citep{mohammadi2024data}) and on a scaled
    Benchmark Simulation Model No.~2 (BSM2) mechanistic
    benchmark~\citep{jeppsson2007bsm2}, whose biological dynamics derive
    from the Activated Sludge Model family (ASM1--ASM3)~\citep{henze2000asm}.
    This combination (two real plants that exhibit opposite failure
    modes of support-only abstention plus a mechanistic oracle)
    evaluates the simulator and the validity layer against both
    observational and counterfactual ground truth in a single protocol.
\end{enumerate}

\section{Background and Related Work}
\label{sec:related}

\subsection{Open-loop simulation versus forecasting}

Most sequence models target forecasting where future inputs are partially
unknown and enter only as side information~\citep{ljung1999sysid,
seborg2017process}.  This framing is misaligned with control-oriented
industrial simulation, where future actuation is specified by design and
must directly drive system
dynamics~\citep{simethy2026ccssrs}.  Transformers with known future
covariates~\citep{lim2021tft} and continuous-time models such as Neural
ODE/CDE~\citep{chen2018neuralode} handle the technical ingredients
(covariates, irregular sampling, partial observation) but are positioned
as general sequence learners rather than open-loop simulators that
explicitly separate planned future actuation from partially observed
state belief.

\subsection{Structured, identifiable and interpretable dynamics}

Classical state-space identification~\citep{ljung1999sysid}, switching
linear dynamical systems~\citep{linderman2017rslds}, and sparse system
identification expose interpretable structure but often struggle under
partial observation, regime switching, or strong nonlinearities.  Two
recent lines sharpen this.  Identifiable recurrent switching dynamical
systems~\citep{balsellsrodas2026rslds} and identifiable latent dynamic
systems~\citep{zhang2025identifiable} place sticky-routed switching on a
representation-learning footing; a parallel line on neural-network
identification of switching nonlinear state-space
models~\citep{zhang2025switchnss} estimates the switching law jointly
with the per-mode dynamics via EM and extended Kalman filtering.
Stable-by-design linear-parameter-varying neural state-space
models~\citep{sertbas2025lpvssm} parameterise context-scheduled
$A(\gamma), B(\gamma)$ with Schur-stability guarantees, and the SOLIS
follow-up~\citep{mansur2026solis} learns physics-informed quasi-LPV
surrogates whose modulation directions are interpretable
physical quantities (natural frequency, damping).  CCSS-IX is closely
related to this LPV-style scheduling, in which each regime $k$ contributes
$A_k(\gamma_t), B_k(\gamma_t), E_k(\gamma_t)$ that adapt with a
low-dimensional context $\gamma_t$.  CCSS-IX adds four load-bearing
differences: (a) regime-switching on a sticky-routed
scaffold~\citep{fox2011sticky}, (b) an explicit disturbance channel
$E_k$ and additive nonlinear response, (c) a probabilistic output head
suited to industrial sensor data (hurdle-Student-$t$), and (d) a
self-falsifying validity layer (\S\ref{sec:validity}) that consumes
the structured decomposition as a counter-example carrier.

Conceptually adjacent: Neural Additive
Models~\citep{agarwal2021nam} learn decomposed response functions
end-to-end; Koopman-with-control~\citep{korda2018koopman,
brunton2022koopman}, SINDy-with-control~\citep{brunton2016sindyc}, and
deep Koopman~\citep{lusch2018koopman} pursue the same
inspectable-update-law goal.  CCSS-IX is closest in mechanism to
FiLM~\citep{perez2018film} and HyperNetwork
parameter-generation~\citep{ha2017hypernet}, restricted to the
controlled state-space form so the conditioning produces
structured $A_k(\gamma), B_k(\gamma), E_k(\gamma)$ rather than
opaque weight tensors.  Structured sequence
models~\citep{gu2022s4} also use input-dependent
state-space parameters but remain general black-box sequence learners
and do not separate state/control/disturbance.  Post-hoc
attribution~\citep{lundberg2017shap} is not a substitute for a
simulator whose update law is itself
decomposable~\citep{rudin2019stop}.

\subsection{Selective prediction, conformal risk control and runtime monitoring}

Selective prediction~\citep{geifman2017selective} and conformal risk
control~\citep{angelopoulos2022crc} trade coverage for risk; sequential
and trajectory-level extensions cover non-exchangeable time
series~\citep{lindemann2023safe} and distribution
shift~\citep{shift2026generative}.  Selective
conformal risk control~\citep{xu2025scrc} more recently integrates
selective classification with conformal risk control, giving a formal
backbone for accept/abstain rules with target-risk guarantees.
Our calibrated-reopen rule (\S\ref{sec:cert}) follows the conformal
risk-control objective of \citet{angelopoulos2022crc} but adds two
domain-specific ingredients, support-novelty gating and
\emph{self-falsification} via event-aligned witnesses, that are
specific to controlled dynamical systems and exploit CCSS-IX's
structured decomposition to produce human-interpretable witnesses
rather than scalar abstention scores.

Pessimism-under-model-uncertainty in offline model-based
RL~\citep{yu2020mopo} plays an analogous role; the validity layer is
also a runtime monitor in the cyber-physical-systems
sense~\citep{cleaveland2024robust}, with safety-monitoring under
distribution shift~\citep{lindemann2025lecps} a closely related
direction.
The most direct neighbour of our temporal-witness mechanism is
optimal-control-based falsification of learnt neural-ODE
surrogates~\citep{falsification2026ocnode}, which produces
signal-temporal-logic counterexamples on a fitted surrogate.  CCSS-IX
differs in two ways: the counterexample carrier is an event-aligned
partition of the same intervention path (inside the operator's
existing vocabulary of aeration changes and influent shifts, rather
than an STL-falsifying perturbation), and attribution decomposes the
disagreement onto the simulator's $A_k$/$B_k$/$E_k$ channels,
reporting \emph{which} channel drove the falsification.

\subsection{Industrial digital twins and wastewater}

Digital twin architectures for industrial processes combine plant data
streams, mechanistic models, learned models, and decision-support
interfaces~\citep{wang2024dt}.  In wastewater
specifically, the recent reviews of \citet{wang2024dt} and
\citet{mehta2025watersector} catalogue an active 2024--2025 literature on DT
deployment in the water sector.  Closer to our N\textsubscript{2}O-aware
framing, \citet{hansen2024lstmn2o} report a black-box long short-term
memory (LSTM) network achieving
$R^2$ up to $\sim$0.98 on 0.5--6\,h N\textsubscript{2}O forecasting from
full-scale WWTP data; \citet{yin2025probwwtp} present a probabilistic
encoder--decoder LSTM for shocking-load effluent forecasting on real
WWTP data; \citet{mdpi2025symmetryn2o} pursue a symmetry-inspired deep
learning approach with explainable post-hoc analysis;
\citet{martinez2026doaeration} pair a foundation model with SHAP for
interpretable dissolved-oxygen forecasting and proactive aeration in
rural plants; \citet{bohn2024denit} study model-class trade-offs
(nonlinear vs.\ linear) under distribution shift on a pilot
denitrification reactor; and \citet{wst2025n2ogann} close a control
loop using a genetic algorithm and neural-network surrogate for
N\textsubscript{2}O minimisation.  Reinforcement-learning controllers
trained against the activated-sludge benchmark
family~\citep{aponte2023rlwwtp} pursue the complementary objective of
policy optimisation; CCSS-IX is positioned upstream as a validity
layer over the learned model itself rather than as a controller.  These works focus on predictive performance, policy
optimisation, or post-hoc explanation of black-box models; none
exposes a decomposed state--control--disturbance update law at the
simulator level, and none provides a self-falsifying decision artefact
for operator screening.

Data-driven WWTP performance reviews~\citep{newhart2019review} document the
distinctive structure of industrial process data (missingness, irregular
phases, nonstationarity, strong temporal dependence) but do not develop
frameworks for auditing learned simulators as selective intervention systems.
The activated-sludge model family ASM1--ASM3~\citep{henze2000asm} provides
the mechanistic foundation for BSM2~\citep{jeppsson2007bsm2}; together with
greenhouse-gas-aware benchmarking~\citep{flores2011ghg} they form the
mechanistic counterpart to our data-driven simulator.
Uncertainty-analysis methodology for WWTP model applications is
established~\citep{sin2009uncertainty}.  The process-control community has
also long studied data-driven monitoring and soft sensing as a way to
recover hard-to-measure variables from instrumented
streams~\citep{qin2012survey,ching2021softsensors}; CCSS-IX inherits this
objective but reframes it as a controlled-simulation-plus-validity problem
rather than a static regression task.

\section{CCSS-IX: Adaptive Structured Simulation for WWTP Dynamics}
\label{sec:ccssix}

\begin{figure*}[!t]
\centering
\includegraphics[width=0.9\linewidth]{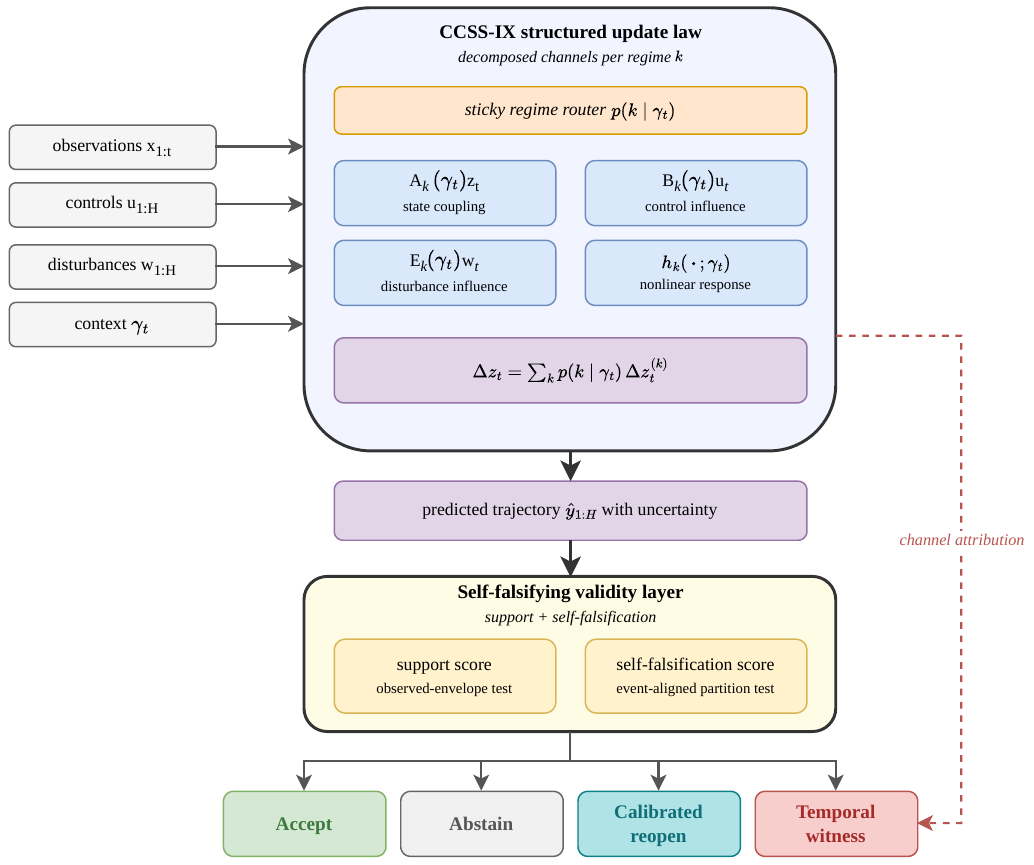}
\caption{System overview.  Plant inputs (observations, planned controls,
disturbances, operating context) feed a sticky regime router; for each
regime $k$ the latent update is the \emph{decomposed} sum of explicit
state-, control-, disturbance-, and nonlinear-response channels
($A_k$, $B_k$, $E_k$, $h_k$), rather than the output of an opaque expert.
The resulting predicted trajectory feeds the self-falsifying validity
layer, which combines a support score with an event-aligned
self-falsification score to issue one of four operator-facing
decisions.  The dashed arrow indicates the use case enabled by the
decomposition: when a temporal witness rejects an apparent approval,
the structured channels support attribution of the disagreement back
to a specific control or disturbance event, turning the validity
outcome into an actionable engineering explanation rather than a scalar
abstention score.}
\label{fig:overview}
\end{figure*}

Fig.~\ref{fig:overview} sketches the overall pipeline.  The CCSS-RS scaffold
\citep{simethy2026ccssrs} provides regime routing, continuous-time
$\Delta t$ handling, and probabilistic output heads.  CCSS-IX replaces the
opaque regime expert with a structured update module whose outputs are
state, control, disturbance, and nonlinear channel contributions.  The
self-falsifying validity layer (\S\ref{sec:validity}) wraps this simulator
into a four-outcome screener.

\subsection{Problem setting: open-loop controlled simulation}
\label{sec:problem}

We consider a partially observed controlled dynamical system with latent
state variables $z_t$, observed outputs $x_t$, known future control inputs
$u_t$, exogenous disturbances $w_t$, irregular time intervals $\Delta t_t$,
and substantial missingness.  The observed outputs $x_t$ are emitted from
$z_t$ through a probabilistic output head (hurdle-Student-$t$, inherited
unchanged from the CCSS-RS scaffold); the structured update law described
below operates in latent space.  The task is to simulate future system
behavior under a specified future control trajectory, conditioned on a
historical context window.

This differs from forecasting in two important ways.  First, the future
control sequence is known and must directly drive the simulator dynamics.
Second, long-horizon fidelity matters more than one-step accuracy, because
prediction errors compound during rollout and can make a simulator unusable
for decision support even if short-horizon metrics are acceptable.

The primary benchmark is the Aved{\o}re N\textsubscript{2}O
dataset~\citep{hansen2024data}, which exhibits all of these properties:
explicit future controls, predominantly 2-min sampling with occasional
gaps, $42.6\%$ overall missingness across state variables, heavy-tailed
state distributions, and evaluation horizons up to 1000 steps.  Two
real plants and the BSM2 oracle (\S\ref{sec:datasets}) provide both
observational stress conditions and a counterfactual ground-truth
slice.

\subsection{Prior work: the CCSS-RS scaffold}
\label{sec:ccssrs}

CCSS-IX builds on CCSS-RS~\citep{simethy2026ccssrs}, a continuous-time
regime-switching scaffold for industrial open-loop simulation.  CCSS-RS
introduced explicit separation of control and state processing, typed
latent initial states, continuous-time rollout encoding with natural
handling of irregular $\Delta t$, sticky regime
routing~\citep{fox2011sticky}, semigroup consistency regularisation, and
probabilistic output heads with Student-$t$ and hurdle distributions
suited to industrial sensor data.  CCSS-RS is used unchanged as the outer
scaffold in this work; the new contribution of CCSS-IX is the replacement
of its opaque black-box regime experts with structured, inspectable
dynamics.

\subsection{Architecture ladder: static and adaptive structure}
\label{sec:static}

Table~\ref{tab:archladder} summarises the architecture ladder: black-box
(CCSS-RS), static structured (CCSS-IX-Static), and adaptive context-conditioned
(CCSS-IX-Adaptive, with and without optional residual).  All variants
operate on the latent state $z_t$ produced by the CCSS-RS scaffold; we
write $z_t, u_t, w_t$ for the latent state, control, and disturbance
inputs to the regime expert at time $t$.  The most direct interpretable
replacement for a black-box expert is a static structured update function
\begin{align}
  f_k(z_t, u_t, w_t) ={}& A_k\,z_t + B_k\,u_t + E_k\,w_t + d_k \notag\\
    & + \textstyle\sum_i \phi_{z_i}(z_{t,i})
      + \textstyle\sum_j \phi_{u_j}(u_{t,j}) \notag\\
    & + \textstyle\sum_m \phi_{w_m}(w_{t,m}),
  \label{eq:static}
\end{align}
optionally with a small residual correction.  This formulation exposes state
coupling matrices $A_k$, control influence matrices $B_k$, disturbance
influence matrices $E_k$, and variable-wise nonlinear response curves
$\phi_{\cdot}$.

A fixed structured parameterisation assumes that a regime is
sufficiently described by a single set of coupling matrices and
response curves.  The 10-seed evaluation in Table~\ref{tab:archladder}
shows that this assumption holds for headline fidelity:
CCSS-IX-Static is statistically indistinguishable from CCSS-RS and
from CCSS-IX-Adaptive on RMSE.  The remaining design question is what
the regime-stratified analyses in \S\ref{sec:res-interp} (channel
attribution, context-modulated response curves, the Causal Isolation
Index) operate against.  A single $A_k$ collapses each regime's
operating-context dependence into one linear map; the adaptive variant
in \S\ref{sec:adaptive} replaces $A_k$ with $A_k(\gamma_t)$ so the
inspectable channels vary with context.  We adopt CCSS-IX-Adaptive as
the main reported model on these grounds, not on fidelity grounds.

\subsection{Adaptive context-conditioned structured dynamics}
\label{sec:adaptive}

CCSS-IX-Adaptive replaces the static structured couplings with
context-conditioned ones, exposing per-regime channel variation that
the static variant cannot.  For each regime $k$, the latent update is
\begin{align}
  \Delta z_t^{(k)} ={}&
    \underbrace{A_k(\gamma_t)\,z_t}_{\text{state contrib.}}
    + \underbrace{B_k(\gamma_t)\,u_t}_{\text{control contrib.}} \notag\\
    &+ \underbrace{E_k(\gamma_t)\,w_t}_{\text{disturbance contrib.}}
    + \underbrace{d_k(\gamma_t)}_{\text{context bias}} \notag\\
    &+ \underbrace{h_k(z_t,u_t,w_t;\,\gamma_t)}_{\text{additive nonlinear}} \notag\\
    &+ \underbrace{\alpha\,r_k(\cdot)}_{\text{optional residual}},
  \label{eq:adaptive}
\end{align}
where $\gamma_t$ is a low-dimensional context vector derived from the
regime embedding, rollout dynamics features, a global context summary, a
compact fast-state summary, and time-step features.  We use the symbol
$h_k$ for the additive nonlinear term in (\ref{eq:adaptive}) to avoid
collision with the variable-wise nonlinear functions $\phi_{\cdot}$ in
(\ref{eq:static}).

The adaptive matrices are parameterised by low-rank modulation:
\begin{equation}
  A_k(\gamma_t) = A_{k,0} + \textstyle\sum_{r=1}^{R} a_{k,r}(\gamma_t)\,\Delta A_{k,r},
  \label{eq:lowrank}
\end{equation}
with analogous forms for $B_k$ and $E_k$.  We use rank $R=4$ in the
main reported model.  This retains an explicit interpretable base
matrix and modulation directions while allowing structured couplings to
adapt across operating contexts.  The additive
nonlinear term preserves a variable-wise decomposition with
context-dependent scale and offset.  In the remainder of the paper we
write $A_k(\gamma)$ without the time subscript when discussing the
parameterisation independently of a specific evaluation time
(figure captions, time-averaged analyses).

Note that $\gamma_t$ does not replace the structured update with an
unrestricted hypernetwork.  It is a compact summary of the current
operating context that selects which structured couplings should be active
at that time.  Structure is preserved; only its parameterisation adapts.
The final main model sets $\alpha=0$: once the structured path is
sufficiently adaptive, the residual is not necessary, and the multi-seed
results confirm this choice on CRPS and NLL
(Table~\ref{tab:archladder}).

\subsection{Inspectable outputs}
\label{sec:inspection}

The parameterisation exposes directly interpretable artifacts without
post-hoc explanation:
\begin{itemize}[leftmargin=2em]
  \item \textbf{State contribution:} coupling among latent state coordinates
    via $A_k(\gamma_t)z_t$;
  \item \textbf{Control contribution:} the effect of planned future actuation
    via $B_k(\gamma_t)u_t$;
  \item \textbf{Disturbance contribution:} exogenous forcing via
    $E_k(\gamma_t)w_t$;
  \item \textbf{Additive nonlinear contribution:} variable-wise shape
    functions with context-dependent scale;
  \item \textbf{Regime occupancy, dwell-time, and transition summaries:}
    directly from the regime-routing layer;
  \item \textbf{Trajectory-level update decomposition:} component shares for
    each rollout window.
\end{itemize}
These are by-products of the simulator's native update law, not secondary
attribution passes over a frozen black-box model.

\section{Self-Falsifying Validity Layer}
\label{sec:validity}

\subsection{Intervention scenarios and decision outcomes}
\label{sec:scenarios}

An intervention scenario consists of recent plant context, a future
control path, disturbance information, a decision threshold (a
water-quality, safety, or risk proxy), and a deployment-style loss
function penalising unsafe accepted scenarios and unnecessary
abstentions~\citep{seborg2017process}.  The validity layer returns one
of four outcomes: \textbf{accept} (passes support and
self-falsification), \textbf{abstain} (unsupported or internally
unstable), \textbf{reopen} (rejected on support, recovered because
self-falsification scores are calibration-normal), or \textbf{witness}
(a raw approval is rejected and returned with an interpretable
event-aligned counterexample).  The four-outcome structure matches
practical industrial screening: a twin should be able to approve,
refuse, recover, or show why a raw approval should not be trusted.

\subsection{Support and self-falsification features}
\label{sec:features}

Write a scenario as $s = (h_{1:t},\,u_{t+1:H},\,w_{t+1:H},\,\tau)$:
context history, planned controls, disturbance forecast, and decision
threshold (a water-quality or safety-proxy bound).  CCSS-IX's raw full
rollout is $\hat{y}(s;\pi_{\text{full}})$ with
$\pi_{\text{full}}=[0,H]$, and the raw decision label is
$d_{\text{raw}}(s)=\mathbf{1}[\max_{t}\hat{y}(s;\pi_{\text{full}})_t < \tau]$.

\textbf{Support features.}  Let $\phi_c(h_{1:t})$ and
$\phi_a(u_{t+1:H})$ be standardised feature embeddings of the recent
context and the planned control path, and let
$\{(\phi_c^{(i)}, \phi_a^{(i)})\}_{i=1}^{N_{\text{cal}}}$ be the
calibration block's $N_{\text{cal}}{=}64$ embedded pairs.  Define
\begin{align}
  \sigma_{\text{ctx}}(s)      &= \tfrac{1}{k}\!\!\sum_{i \in \mathcal{N}_k(\phi_c)} \!\!\|\phi_c - \phi_c^{(i)}\|_2, \notag\\
  \sigma_{\text{act}|\text{ctx}}(s) &= \tfrac{1}{k}\!\!\sum_{i \in \mathcal{N}_k(\phi_c)} \!\!\|\phi_a - \phi_a^{(i)}\|_2, \notag\\
  \sigma_{\text{supp}}(s)     &= 0.35\,\sigma_{\text{ctx}}(s) + 0.65\,\sigma_{\text{act}|\text{ctx}}(s),
  \label{eq:support}
\end{align}
where $\mathcal{N}_k(\phi_c)$ is the index set of the $k=10$
calibration neighbours closest to $\phi_c$ in Euclidean distance, and
the action distance is measured to the actions of those context
neighbours (so the score is action-conditional on the context).  The
weights $(0.35, 0.65)$ tilt the score toward action-conditional
novelty (the operationally critical axis when the recent context is
well represented but the proposed control path drifts away from
historical operation), and are held fixed across plants rather than
retuned per dataset.  A joint context+action $k$-NN distance is
reported as an auxiliary diagnostic.  A scenario is \emph{supported} iff $\sigma_{\text{supp}}(s)
\le q_\alpha$, where $q_\alpha$ is the $(1-\alpha)$-quantile of
$\sigma_{\text{supp}}$ over the calibration block at $\alpha=0.10$
\citep{angelopoulos2022crc}.
Support-only abstention is an operationally-plausible baseline (an
engineer may decide not to trust the twin far from observed
operation), but the paper's hypothesis is that support alone is
insufficient: some unsafe interventions remain supported and some safe
interventions are needlessly rejected as novel.

\textbf{Self-falsification features.}  For a partition family
$\mathcal{P}$ over $[0,H]$, the \emph{partition orbit} is
$\mathcal{O}(s,\mathcal{P}) = \{\hat{y}(s;\pi) : \pi \in
\mathcal{P}\}$, where $\hat{y}(s;\pi)$ is the rollout under partition
$\pi=(0,t_1,\ldots,t_{J-1},H)$ recomposed segment-by-segment from
CCSS-IX.  Let $m(s,\pi;\tau) = \tau - \max_t \hat{y}(s;\pi)_t$ be the
per-partition margin.  The decision-facing \emph{partition defect}
\begin{equation}
  \phi_{\text{self}}(s,\mathcal{P};\tau) \;=\;
    \mathbf{1}\!\bigl[\,\min_{\pi\in\mathcal{P}} m(s,\pi;\tau) \,<\, 0
                  \,\le\, \max_{\pi\in\mathcal{P}} m(s,\pi;\tau)\bigr]
  \label{eq:defect}
\end{equation}
is the orbit-straddles-threshold indicator: at least one decomposition
predicts unsafe while at least one (including the raw full rollout)
predicts safe.  The auxiliary scalar
$\phi_{\text{orbit}}(s,\mathcal{P}) = \max_{\pi\in\mathcal{P}}
\max_t \hat{y}(s;\pi)_t - \min_{\pi\in\mathcal{P}} \max_t
\hat{y}(s;\pi)_t$ is the orbit diameter on the peak statistic; we use
$\phi_{\text{self}}$ for decisions and $\phi_{\text{orbit}}$ as a
reportable diagnostic.

\textbf{Event-aligned cut selection.}  We construct
$\mathcal{P}_{\text{eva}}$ from the controls and disturbances directly.
Normalise the joint signal $\nu_t = (u_t, w_t)$ by its per-feature
median absolute deviation $\hat{s}_j = \mathrm{MAD}(\nu_{\cdot,j})$,
form the step-wise norm $\Delta_t = \|\nu_{t+1}/\hat{s} -
\nu_t/\hat{s}\|_2$, and pick the top-$k$ steps:
\begin{equation}
  \mathcal{C}_k(s) \;=\; \arg\!\mathrm{top}\,k\,\bigl\{\Delta_t :
                                                       t=1,\ldots,H-1\bigr\}.
  \label{eq:cuts}
\end{equation}
Event-aligned B4 takes $k=3$ and assembles a partition family
$\mathcal{P}_{\text{eva}}$ with $1$, $2$, and $3$ interior cuts
drawn from $\mathcal{C}_3(s)$; combined with the raw full rollout in
the partition-defect test (Eq.~\ref{eq:defect}), these form the
four-element comparison orbit that gives the family its name
(``B4'').  This is a data-driven cut family that places candidate
decompositions at the operationally-meaningful events the operator
would already flag (aeration changes, influent shifts, wet-weather
pulses).

\begin{figure*}[!t]
\centering
\includegraphics[width=1\linewidth]{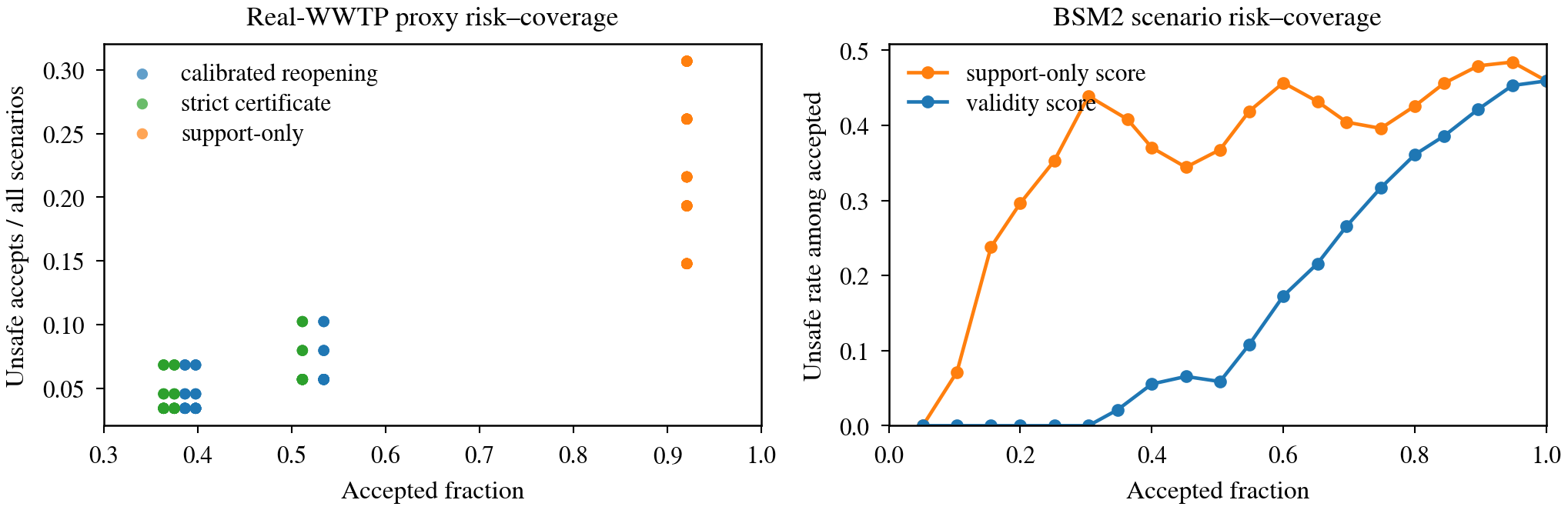}
\caption{Risk--coverage view of the validity layer on the two-plant
shadow-mode replay.  Support-only abstention (blue) trades coverage for
risk along a curve dominated by the calibrated-reopen policy (orange) at
moderate-to-high unsafe-cost weights.  Strict certification (green) lies
between them, eliminating unsafe acceptance at the cost of discarding
calibration-normal recoverable scenarios.}
\label{fig:risk-coverage}
\end{figure*}

\subsection{Strict certificate, calibrated reopening, and event-aligned witnesses}
\label{sec:cert}

The \textbf{strict certificate} accepts $s$ only when $s$ is supported
AND the orbit does not straddle the threshold.  The \textbf{calibrated
reopen} rule reopens an unsupported scenario when its self-falsification
scores fall in the calibration-normal region
\begin{equation*}
  \mathcal{C}_\alpha \;=\; \bigl\{(\sigma_{\text{supp}}, \phi_{\text{orbit}}) :
    \sigma_{\text{supp}} \le q^{\text{supp}}_{1-\alpha},\,
    \phi_{\text{orbit}} \le q^{\text{orbit}}_{1-\alpha}\bigr\},
\end{equation*}
the product of the marginal $(1-\alpha)$-quantile bounds of each score over
the calibration block, instantiating \citet{angelopoulos2022crc}'s
conformal-risk-control objective on the structured-decomposition scores.  A
\textbf{temporal witness} is issued when the raw rollout accepts but
event-aligned decomposition produces a falsifying partition; the
returned witness is the most-unsafe partition with channel attribution
back to CCSS-IX's $A_k(\gamma_t), B_k(\gamma_t), E_k(\gamma_t)$ updates.

Alg.~\ref{alg:validity} pins down the four-outcome rule.  Compared to
raw-full hard gating, the event-aligned family is the primary paper
object because its witnesses are describable in engineering terms (the
same low-DO scenario that passes full rollout fails when the path is
evaluated in segments aligned with aeration changes), and because the
simulator's exposed channels let the engineer inspect which channel
drove the disagreement.  We also report three contrast families for comparison: \emph{dyadic}
decomposition (a single mid-horizon cut, splitting the rollout in two);
\emph{matched-random B4} (the same four-element orbit form but with
interior cuts drawn uniformly at random rather than from
$\mathcal{C}_3(s)$); and \emph{decision-stressed B4} (interior cuts
chosen to maximally stress model margins, providing a detection upper
bound).

\begin{algorithm}[t]
  \caption{CCSS-IX self-falsifying validity layer.}
  \label{alg:validity}
  \begin{algorithmic}[1]
    \Require Scenario $s=(h_{1:t}, u_{t+1:H}, w_{t+1:H}, \tau)$;
             calibration block with support quantile $q_\alpha$ and
             calibration-normal region $\mathcal{C}_\alpha$
    \Ensure Outcome $\in \{\textsc{accept}, \textsc{abstain},
             \textsc{reopen}\}$ or a \textsc{witness} pair $(\pi^*, \xi^*)$
             of partition and channel-attribution payload
    \State $\hat{y}_{\text{full}} \gets \text{CCSS-IX}(s; \pi_{\text{full}})$
            \Comment{raw rollout}
    \State $d_{\text{raw}} \gets \mathbf{1}[\max_t \hat{y}_{\text{full},t} < \tau]$
            \Comment{raw safe label}
    \State $\sigma_{\text{supp}} \gets$ support score from
            Eq.~\eqref{eq:support}
    \State $\mathcal{C}_3 \gets$ event-aligned cuts from
            Eq.~\eqref{eq:cuts}
    \State $\mathcal{P}_{\text{eva}} \gets$ B4 partitions of
            $[0,H]$ built from $\mathcal{C}_3$ (1, 2, and 3 interior cuts)
    \State $\phi_{\text{self}} \gets$ partition defect from
            Eq.~\eqref{eq:defect} on
            $\mathcal{P}_{\text{eva}} \cup \{\pi_{\text{full}}\}$
    \State $\phi_{\text{orbit}} \gets \max_{\pi}\max_t \hat{y}(s;\pi)_t
            - \min_{\pi}\max_t \hat{y}(s;\pi)_t$
            \Comment{orbit diameter on peak}
    \If{$\sigma_{\text{supp}} > q_\alpha$} \Comment{unsupported}
        \If{$(\sigma_{\text{supp}}, \phi_{\text{orbit}})
             \in \mathcal{C}_\alpha$}
            \State \Return \textsc{reopen}
        \Else
            \State \Return \textsc{abstain}
        \EndIf
    \EndIf
    \If{$d_{\text{raw}} = 1$ \textbf{and}
        $\phi_{\text{self}} = 1$} \Comment{witness exists}
        \State $\pi^* \gets
                \arg\min_{\pi \in \mathcal{P}_{\text{eva}}}
                m(s, \pi; \tau)$ \Comment{most-unsafe cut}
        \State $k^* \gets$ regime occupied at
                $\arg\max_t \hat{y}(s;\pi^*)_t$
        \State $\xi^* \gets
                \bigl( A_{k^*}(\gamma_t)z_t,\, B_{k^*}(\gamma_t)u_t,\,
                       E_{k^*}(\gamma_t)w_t \bigr)$ at witness peak
        \State \Return $(\pi^*, \xi^*)$ as \textsc{witness}
    \Else
        \State \Return \textsc{accept}
    \EndIf
  \end{algorithmic}
\end{algorithm}

\textbf{Worked example (low-DO scenario, BSM2).}  Consider the
representative scenario in Fig.~\ref{fig:witness} (BSM2 scenario library
identifier low-DO strategy~04).  CCSS-IX's raw full rollout peaks at
$\hat{y}=0.0426$, below the N\textsubscript{2}O threshold
$\tau=0.0465$, so $d_{\text{raw}}=1$ (raw safe label).  The scenario is
supported ($\sigma_{\text{supp}} \le q_\alpha$), so the algorithm
proceeds to the witness check.  Event-aligned cut selection
(Eq.~\ref{eq:cuts}) ranks the aeration step at $t=6$ as the top
control discontinuity, producing partition $\pi^* = [0,6,16]$ whose
recomposed rollout peaks at $0.0549 > \tau$, so $\phi_{\text{self}}=1$
(Eq.~\ref{eq:defect}).  The algorithm returns
\textsc{witness}$(\pi^*, \cdot)$; CCSS-IX's $B_k(\gamma_t)u_t$ channel
attributes the disagreement to the aeration set-point step at $t=6$.
The BSM2 oracle confirms the witness was correct: the true peak is
$0.0599$, also above $\tau$.  The four-outcome layer thus converted a
false-safe approval into a mechanism-attributed refusal. Fig.~\ref{fig:risk-coverage} previews the resulting risk--coverage
trade-off on the two-plant shadow-mode replay; the per-plant
breakdown is reported in \S\ref{sec:res-wwtp}.

\begin{table*}[!t]
\centering
\caption{Architecture ladder on the Aved{\o}re primary benchmark.
Multi-seed results for the CCSS family report mean $\pm$ std across
\textbf{ten seeds} $\{1,5,7,11,19,23,31,42,47,53\}$; each row first
averages over the five state variables (NH4, NO3, N\textsubscript{2}O,
O2, SS), then over seeds.  ``Paired diff vs CCSS-RS'' is the mean
within-seed RMSE difference, with 95\% bootstrap CI from $20{,}000$
resamples of the paired differences.  $\Delta$RMSE (\%) is the mean of
per-seed ratios $(\text{model}_i - \text{CCSS-RS}_i)/\text{CCSS-RS}_i$.
All CCSS-IX variants share the CCSS-RS outer scaffold; only the regime
expert differs.  At 10 seeds, each paired bootstrap CI on RMSE crosses
zero (Static, Adaptive, Adaptive+Res).  Lower is better; bold marks
the main reported model.  See \S\ref{sec:res-arch} for the variant
selection rationale.}
\label{tab:archladder}
\setlength{\tabcolsep}{5pt}
\footnotesize
\begin{tabular}{lccccc}
\toprule
& \multicolumn{3}{c}{Multi-seed (mean $\pm$ std)} & \multicolumn{2}{c}{Paired diff vs CCSS-RS} \\
\cmidrule(lr){2-4}\cmidrule(lr){5-6}
Model & RMSE & CRPS & NLL & $\Delta$RMSE (\%) & 95\% bootstrap CI \\
\midrule
LSTM open-loop (adapted)$^{\dagger}$ & $1.158 \pm 0.137$ & $0.789 \pm 0.114$ & --- & $+0.535 \pm 0.163$ (+86.8\%) & $[+0.390,\,+0.685]$ \\
CCSS-RS (black-box reference, $n{=}10$) & $0.6435 \pm 0.045$ & $0.3825 \pm 0.041$ & $0.7784 \pm 0.339$ & --- & --- \\
CCSS-IX-Static ($n{=}10$)              & $0.6459 \pm 0.040$ & $0.3830 \pm 0.028$ & $0.7547 \pm 0.154$ & $+0.002 \pm 0.058$ (+0.78\%) & $[-0.029,\,+0.038]$ \\
\textbf{CCSS-IX-Adaptive} (main, $n{=}10$) & $\mathbf{0.6502 \pm 0.051}$ & $\mathbf{0.3900 \pm 0.040}$ & $\mathbf{0.8201 \pm 0.248}$ & $\mathbf{+0.007 \pm 0.030}$ (+1.08\%) & $[-0.010,\,+0.024]$ \\
CCSS-IX-Adaptive+Res ($n{=}10$)        & $0.6644 \pm 0.061$ & $0.4215 \pm 0.049$ & $1.2848 \pm 0.368$ & $+0.021 \pm 0.051$ (+3.36\%) & $[-0.010,\,+0.050]$ \\
\bottomrule
\end{tabular}
\par\vspace{4pt}
\begin{minipage}{\linewidth}
  \scriptsize
  $^{\dagger}$Adapted LSTM at matched 1.21\,M parameter budget with
  mask-aware Gaussian NLL and log1p$(\Delta t)$-conditioned inputs to
  handle 43\% missingness and irregular sampling: the same adaptations
  the underlying CCSS-RS preprint \citep{simethy2026ccssrs} considers as
  taking recurrent baselines outside the like-for-like regime. NLL omitted
  (Gaussian head vs CCSS hurdle-Student-t is not directly comparable). The
  LSTM row aggregates four seeds $\{1,7,19,42\}$; its paired diff vs
  CCSS-RS uses those same four seeds for matched comparison, while the
  CCSS family rows aggregate the full 10-seed set. The row quantifies the
  cost of LSTM adaptations rather than claiming a head-to-head; we use
  Agtrup (regular grid, no missingness) for the like-for-like recurrent
  and SSM comparison.
\end{minipage}
\end{table*}

\section{Experimental Evaluation}
\label{sec:experiments}

\subsection{Datasets and experimental design}
\label{sec:datasets}

\noindent\textbf{Aved{\o}re WWTP (primary benchmark).}
The Aved{\o}re N\textsubscript{2}O dataset~\citep{hansen2024data} covers
two years (June 2022 -- June 2024) of a full-scale municipal wastewater
treatment plant serving 350{,}000 population equivalents (PE) near Copenhagen, Denmark.  The
dataset contains 906{,}815 timesteps with 16 variables from Tank~1 across
five categories: state variables (NH\textsubscript{4},
NO\textsubscript{3}, N\textsubscript{2}O, O\textsubscript{2}, SS), known
future controls (6 valve and setpoint variables), exogenous disturbances
(5 flow, airflow, temperature variables), and categorical process phase
encodings.  Key challenges include $42.6\%$ overall missingness across
state variables (uniform across NH$_4$, NO$_3$, N$_2$O, O$_2$, and SS),
predominantly 2-min sampling with rare multi-minute gaps and
occasional multi-hour outages, $35.2\%$ of valid N$_2$O readings near
zero (below the $0.001$ hurdle threshold), and heavy-tailed state
variables.  An LSTM-based N\textsubscript{2}O forecasting baseline on
comparable full-scale WWTP data is reported by \citet{hansen2024lstmn2o};
our task is
different (open-loop simulation under specified future controls, with
event-aligned witness generation) and uses a different evaluation
protocol (locked chronological shadow-mode replay).

\noindent\textbf{Agtrup WWTP (transfer validation).}
The Agtrup/BlueKolding nutrient-removal
dataset~\citep{mohammadi2024data} provides a second full-scale plant with
similar variable structure but a distinct operational regime and focus
on nutrient removal rather than N\textsubscript{2}O dynamics.  It serves
two purposes: (i)~a matched-budget multi-site head-to-head against LSTM
and S5 baselines that cannot run natively on Aved{\o}re
(\S\ref{sec:res-agtrup}); and (ii)~exposure of the
\emph{safe-unsupported-recovery} failure mode of support-only
abstention in the shadow-mode replay (\S\ref{sec:res-wwtp}).  Each plant
is trained, calibrated, and evaluated on its own chronological splits;
we do not perform a zero-shot Aved{\o}re$\to$Agtrup transfer.

\noindent\textbf{BSM2 (mechanistic oracle).}
Benchmark Simulation Model No.~2~\citep{jeppsson2007bsm2} provides a
whole-plant mechanistic wastewater simulator built on the
activated-sludge model family~\citep{henze2000asm} and including
clarification, sludge treatment, and anaerobic digestion.  BSM2 is used
here as a mechanistic oracle for counterfactual evaluation: intervention
scenarios are generated and evaluated against the mechanistic model where
the ground-truth response is known.  Greenhouse-gas-aware extensions of
BSM2~\citep{flores2011ghg} motivate the energy--N\textsubscript{2}O
frontier explored in \S\ref{sec:res-witness}.

\noindent\textbf{Calibration protocol.}
For all real-plant evaluations, we separate the available chronological
record into (i) a training block, (ii) a calibration block disjoint
from both training and evaluation (used to compute the support quantile
$q_\alpha$ and the self-falsification calibration-normal region
$\mathcal{C}_\alpha$), and (iii) a frozen held-out evaluation block in
chronological order.  The Aved{\o}re validity-layer calibration block
contains $N_{\text{cal}}=64$ windows (matched score-train and
evaluation block sizes); BSM2 uses 8 calibration episodes $\times$ 12
windows per seed for the witness-matrix run.  The validity layer
writes decisions before outcome scoring; observed target windows are
used afterward to compute unsafe acceptance and regret.  Headline
comparisons use threshold quantile 0.90 (the $1{-}\alpha$ quantile rule
of \citet{angelopoulos2022crc} with target risk $\alpha = 0.10$) and
unsafe-cost weight 4 unless otherwise stated.  All headline analyses are
frozen; no threshold is tuned on held-out evaluation rows after canonical
rows are selected.  We caution that
calibration-block representativeness is a limitation when distribution
shift, sensor drift, or process changes occur, as noted in
\S\ref{sec:limits}.

\noindent\textbf{Statistical reporting.}
Reported point metrics are: root-mean-square error (RMSE) and mean
absolute error (MAE) for trajectory accuracy; coefficient of determination
($R^2$); continuous ranked probability score (CRPS) and negative
log-likelihood (NLL) for probabilistic calibration.  Lower is better for
all except $R^2$.  Tables~\ref{tab:archladder}--\ref{tab:witnesses} pair
these point estimates with confidence intervals where appropriate.
For paired binary outcomes on the same evaluation pool (witness matrix),
we report a McNemar exact test on the worst-case (maximum-concordance)
discordance under the marginal counts; for small-$n$ proportions (BSM2
main slice with $n=3$), we report a 95\% Clopper--Pearson interval; for
two-plant regret comparisons, the per-plant breakdown in
Table~\ref{tab:shadow} exposes the within-group sample sizes that drive
the aggregate.

\noindent\textbf{Multi-seed protocol.}
All four rungs of the CCSS architecture ladder (CCSS-RS,
CCSS-IX-Static, CCSS-IX-Adaptive, CCSS-IX-Adaptive+Res) are
multi-seeded over ten random seeds
$\{1, 5, 7, 11, 19, 23, 31, 42, 47, 53\}$.  Each seed re-initialises
the model weights and the random window sampler; data splits,
calibration windows, and all other pipeline state remain frozen across
seeds.  Aggregates in Table~\ref{tab:archladder} are reported as mean
$\pm$ standard deviation across seeds, computed after averaging within
each seed over the five Aved{\o}re state variables.  Paired comparisons
use the within-seed RMSE difference on the shared seeds, with 95\%
bootstrap confidence intervals computed from 20{,}000 resamples of the
paired differences.  The LSTM baseline (Table~\ref{tab:archladder} top
row) is reported on the original four-seed subset $\{1,7,19,42\}$ for
matched paired comparison against CCSS-RS on those seeds.

\subsection{Architecture ladder under matched seeds}
\label{sec:res-arch}

Table~\ref{tab:archladder} reports the architecture ladder on the
Aved{\o}re benchmark under a matched ten-seed protocol (seeds
$\{1,5,7,11,19,23,31,42,47,53\}$).  Replacing the CCSS-RS black-box
regime expert with structured experts (static, adaptive
context-conditioned, or adaptive+residual) yields RMSE values that
are statistically indistinguishable from the black-box reference:
every paired bootstrap CI vs CCSS-RS crosses zero
(Table~\ref{tab:archladder}, Static $+0.78\%$, Adaptive $+1.08\%$,
Adaptive+Res $+3.36\%$).  CRPS values cluster inside within-seed
variance; CCSS-IX-Adaptive trails CCSS-RS marginally on probabilistic
calibration, while the static variant ties CCSS-RS on CRPS and edges
it on NLL.  We retain CCSS-IX-Adaptive as the main model because its
regime-stratified context-modulated couplings $A_k(\gamma_t),
B_k(\gamma_t), E_k(\gamma_t)$ are the substrate for the interpretability
and witness analyses in \S\ref{sec:res-interp} onwards, even
though the static variant is competitive on the headline fidelity
metrics.  CCSS-IX-Adaptive+Res is dominated on CRPS and NLL, so the
adaptive variant's choice $\alpha=0$ is empirically vindicated.

Per-seed Static-vs-CCSS-RS ratios span $-9\%$ to $+17\%$, wide
enough that pre-registering an evaluation seed set is necessary for
any headline claim on this benchmark.

\noindent\textbf{Adapted-LSTM reference.}
For completeness, we also report an LSTM open-loop baseline at matched
1.21\,M parameter budget on the same four seeds (Table~\ref{tab:archladder},
top row).  The LSTM is given mask-aware Gaussian NLL and
log1p$(\Delta t)$-conditioned inputs to operate on Aved{\o}re's irregular
sampling and 43\% missingness; the underlying CCSS-RS
preprint~\citep{simethy2026ccssrs} argues that these adaptations push
recurrent baselines outside the like-for-like regime.  Reported here as a \emph{cost of
adaptation}: even with those modifications, the adapted LSTM trails
CCSS-RS by $86.8\%$ paired RMSE (95\% bootstrap CI $[+0.390,+0.685]$),
materially behind every rung of the CCSS-IX family.  The clean
recurrent/SSM head-to-head (LSTM, Mamba, S5 against the CCSS family)
is run on the Agtrup dataset (regular 2-min grid, $0\%$ missingness)
in \S\ref{sec:res-agtrup}, where those baselines apply natively.

\begin{figure*}[!b]
\centering
\includegraphics[width=0.95\linewidth]{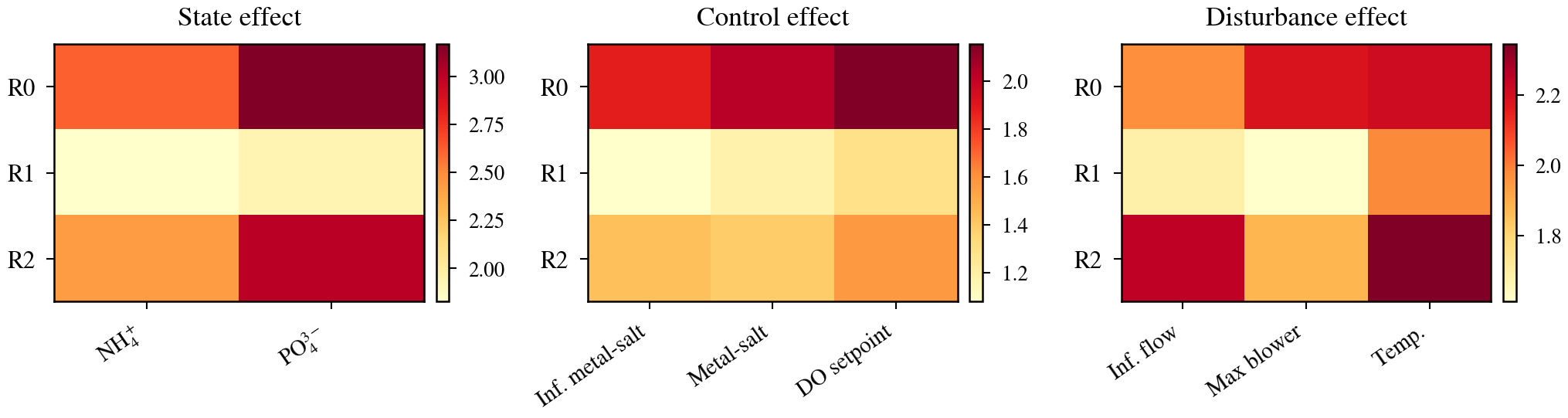}
\caption{Channel-effect strength per regime at Agtrup (state /
control / disturbance, left to right).  Each cell is the total
contribution of one variable's channel ($A_k(\gamma)$, $B_k(\gamma)$,
$E_k(\gamma)$) under the held-out evaluation window; rows are
regimes 0--2.  The numerical values are reported in the supplementary
material as per-channel effect tables.}
\label{fig:channel-effects}
\end{figure*}

\subsection{CCSS-IX interpretability: regimes, response curves, and N\textsubscript{2}O attribution}
\label{sec:res-interp}

\noindent\textbf{Regime structure (Aved{\o}re primary benchmark).}
The learned regimes are persistent and parametrically distinct, not
routing artefacts.  Three regimes are recovered with balanced
occupancies (0.22, 0.57, 0.22) and mean dwell times of 16.5, 47.9,
and 10.9 steps, the persistence pattern of sticky
routing~\citep{fox2011sticky}.

\begin{figure*}[t]
\centering
\includegraphics[width=0.85\linewidth]{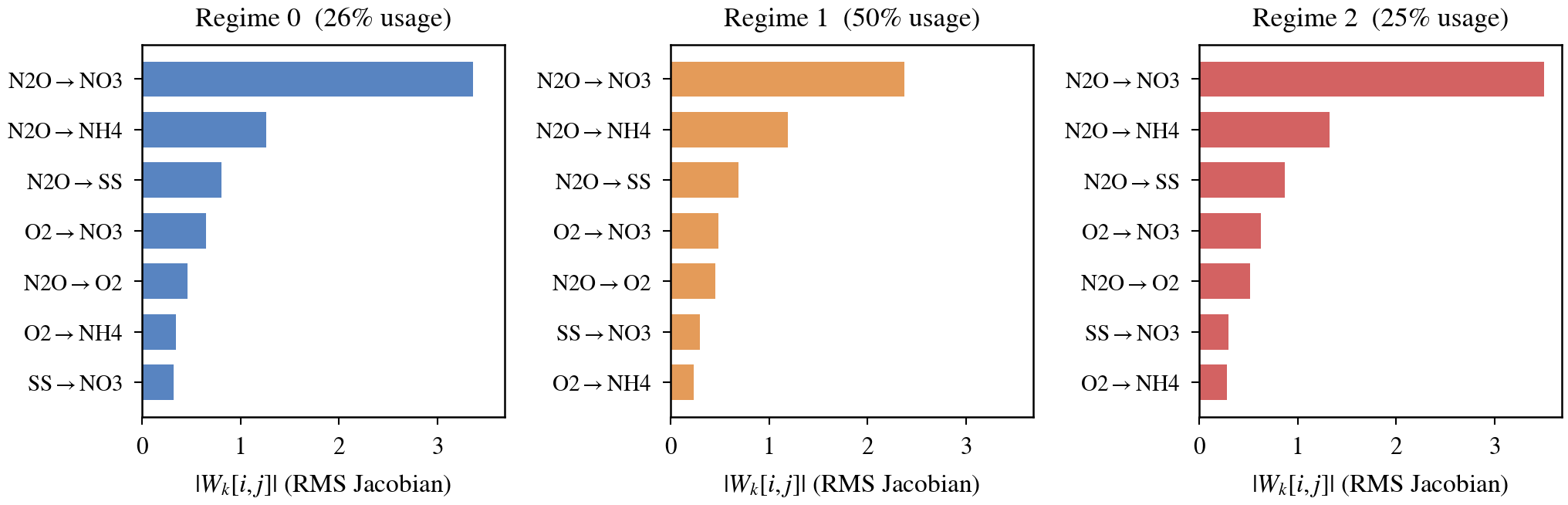}
\caption{Top dependency edges per regime ranked by $|W_k[i,j]|$ on the
held-out Aved{\o}re evaluation block.  The dominant edges
(N$_2$O-driven couplings into NO$_3$, NH$_4$, and SS, plus
O$_2$$\to$NO$_3$) are shared across all three regimes
(Spearman rank correlation of edge magnitudes $\geq 0.98$ for every
pairwise regime comparison), so the regime-specific signal is carried by
edge \emph{magnitude} rather than topological rewiring.  Regime~1 is
the lower-magnitude (``quiescent'') regime; regimes~0 and~2 are roughly
$22$--$25\%$ higher in aggregate Jacobian norm.  This is consistent
with regimes acting as activity-level gates on universal biological
coupling rather than switching between qualitatively different
reaction networks.  The full $A_k$ matrices in heat-map form are
reported in the supplementary material.}
\label{fig:edges}
\end{figure*}

The three regimes differ substantially in their per-channel coupling
weights: the mean pairwise Frobenius-norm distance between the
vectorised regime coupling maps is $2.53$, on the order of the
matrices' own entry-magnitude scale rather than within numerical
noise.  Their aggregate Jacobians, however, share the same nonzero-edge
pattern and differ mainly in magnitude (Fig.~\ref{fig:edges}).  The
regimes therefore act as activity-level gates over a single shared
biological coupling structure: different rates and intensities of the
same underlying chemistry, not different chemistries.

To check whether these regimes reflect physically meaningful structure,
we compare each regime assignment against two known operational
signals (the storm-water inflow indicator and the configured
process-phase setpoint) by computing Cram\'er's $V$ between the
per-window modal regime and the per-window modal value of each signal.
Storm-water inflow tracks the regime closely ($V \approx 0.91$), and
the phase setpoint moderately ($V \approx 0.78$).  The regimes
therefore reflect real operating-context variation without reducing to
either named signal alone; they decompose plant operation along an
activity-level axis jointly defined by inflow and aeration, which is
coarser than the named-phase partition.

\noindent\textbf{Synthetic-system validation.}
On a controlled 2-D switching system with a known LPV ground-truth update
law and irregular sampling (Fig.~S4 in the supplementary material), static
structured experts roll out at RMSE $0.583$ while the adaptive
context-conditioned variant rolls out at RMSE $0.089$, a
$6.5{\times}$ reduction on a system where the LPV scheduling form is
the right inductive bias.  The adaptive LPV per-regime law recovers
the ground-truth law that a fixed structured parameterisation cannot.

\noindent\textbf{Coupling matrices and ASM-family mechanistic priors.}
The literal $A_k$ coupling matrices report the unsummed update law
applied at inference time; Fig.~\ref{fig:edges} renders the implied
regime-stratified dependency graph as the top edges per regime.
Table~\ref{tab:asm1} compares CCSS-IX against a literature-prior edge
list combining ASM1 nitrification/denitrification
relations~\citep{henze2000asm} with N$_2$O production pathways from
the wastewater N$_2$O-mechanism literature~\citep{kampschreur2009n2o}
(standard ASM1 does not model N$_2$O).  Six of eight edges match at
the $5\%$-of-row-maximum threshold; the two non-matches
($\text{N}_2\text{O}\!\leftarrow\!\text{NO}_3$ at $4.0\%$,
$\text{O}_2\!\leftarrow\!\text{NH}_4$ at $4.2\%$) sit within
$1\,\text{pp}$ of the threshold and are recovered at slightly higher
cutoffs (Fig.~S1).  The structured channels recover the qualitative
mechanistic shape \emph{without ever being fitted to mechanistic-model
outputs}: emergent agreement between a learned structured law and
the activated-sludge modelling tradition.

\begin{table*}[t]
\centering
\caption{Literature-prior mechanistic-edge recovery from CCSS-IX's
learned coupling matrices on the held-out Aved{\o}re evaluation block.
Eight edges are queried (ASM1 nitrification/denitrification process
relations~\citep{henze2000asm}; N$_2$O-bearing edges follow the
wastewater N$_2$O-mechanism literature~\citep{kampschreur2009n2o} since
standard ASM1 does not model N$_2$O).  CCSS-IX matches six at the
$5\%$-of-row-maximum threshold without ever being trained against
mechanistic-model outputs; the two non-matches sit within
$1\,\text{pp}$ of the threshold and are recovered at slightly higher
thresholds (Fig.~S1).  Edge weight $W_{\text{global}}$ aggregates the
regime-stratified coupling magnitudes; ``\% of row max'' normalises
within the affected state.}
\label{tab:asm1}
\setlength{\tabcolsep}{8pt}
\small
\begin{tabular}{llcccc}
\toprule
Effect on & Cause & Regime context & $W_{\text{global}}$ & \% of row max & ASM1 match \\
\midrule
NH$_4$ & O$_2$  & aerobic & 0.2771 & 22.3\,\% & \checkmark \\
NO$_3$ & O$_2$  & aerobic & 0.5605 & 19.2\,\% & \checkmark \\
NH$_4$ & NO$_3$ & anoxic  & 0.1020 & \phantom{0}8.2\,\% & \checkmark \\
N$_2$O & O$_2$  & anoxic  & 0.0716 & 12.3\,\% & \checkmark \\
N$_2$O & NO$_3$ & anoxic  & 0.0234 & \phantom{0}4.0\,\% & $\times$ \\
O$_2$  & NH$_4$ & aerobic & 0.0198 & \phantom{0}4.2\,\% & $\times$ \\
NO$_3$ & NH$_4$ & anoxic  & 0.1673 & \phantom{0}5.7\,\% & \checkmark \\
N$_2$O & N$_2$O & both    & 0.5819 & 100.0\,\% & \checkmark \\
\bottomrule
\end{tabular}
\end{table*}

\noindent\textbf{Eigenmode time-scales and threshold-sensitivity.}
Eigendecomposition of $A_k(\gamma)$ recovers regime-specific
time-constants per state variable (Fig.~S5; range $\sim$4--97 min
across regimes), and a threshold-sweep over the supported calibration
range (Fig.~S1) shows the model ranking is stable, so the
$1{-}\alpha=0.90$ rule is not knife-edged.

\noindent\textbf{Channel-effect heatmaps (Agtrup).}
On the Agtrup model, summing $A_k(\gamma), B_k(\gamma), E_k(\gamma)$
per regime over a held-out window gives a one-glance per-channel
ranking (Fig.~\ref{fig:channel-effects}).  Three patterns are stable.
Tank-1 phosphate (PO$_4^{3-}$) and ammonium (NH$_4^+$) dominate the
state side (per-cell totals $1.83$--$3.17$), and the context-dependent
modulation dominates the base coupling by roughly three-to-one in
operator norm ($\|\sum_r a_{k,r}(\gamma_t)\,\Delta A_{k,r}\|$ vs.\
$\|A_{k,0}\|$, averaged over the held-out window).  This is the structural
reason the adaptive variant exposes inspectable context-dependence
even though the static variant matches its headline fidelity
(\S\ref{sec:res-arch}).  Dissolved-oxygen
set-point and metal-salt dosing (including the influent metal-salt
stream) dominate controls in every regime (totals $1.0$--$2.2$), matching
operator domain knowledge of the load-bearing levers at Agtrup.
Finally, the inference-time coefficient magnitudes order as
disturbance $\mathbb{E}|E|=6.24 >$ state $2.85 >$ control $0.46$: at
Agtrup the disturbances move the latent state more than the controls
do.  Each heatmap cell is a number a process engineer can rank and
falsify: the structured decomposition is the model's update law,
not a wrapper over a black box.

\noindent\textbf{Response curves (Agtrup).}
Each per-variable nonlinear contribution $\phi_x, \phi_u, \phi_w$ is a
one-dimensional function that the model can be queried for directly
(highlights from the Agtrup model in Fig.~S6 of the supplementary
material).  Tank-1 phosphate (PO$_4^{3-}$) ranks highest by curve-shape
importance (selection score $0.238$) with a saturating non-linearity
consistent with nutrient-pool self-buffering; metal-salt dose dominates
controls (score $0.276$) with the sub-linear shape typical of
phosphorus precipitation chemistry; influent flow (score $0.114$)
shows a clear non-linear ramp, while maximum blower capacity (score
$0.097$) and process temperature (score $0.076$) are more nearly
linear.  Each curve is the literal additive contribution evaluated at
inference time, not a post-hoc visualisation: an operator who disagrees
with the PO$_4^{3-}$ curve has named, replaceable disagreement with a
specific component of the update law.

\noindent\textbf{Regime case-study trajectories (Agtrup).}
Three Agtrup windows, one per target regime, decomposed into the five
channels (Fig.~S7), exhibit regime-dependent channel balance:
disturbance-dominant in regimes~0 and~2, control-dominant in regime~1
(consistent with $23/24$ N\textsubscript{2}O spikes sitting in regime~1
on the Aved{\o}re benchmark below).  Residual share is zero in every
window.

\begin{figure*}[!t]
\centering
\includegraphics[width=1\linewidth]{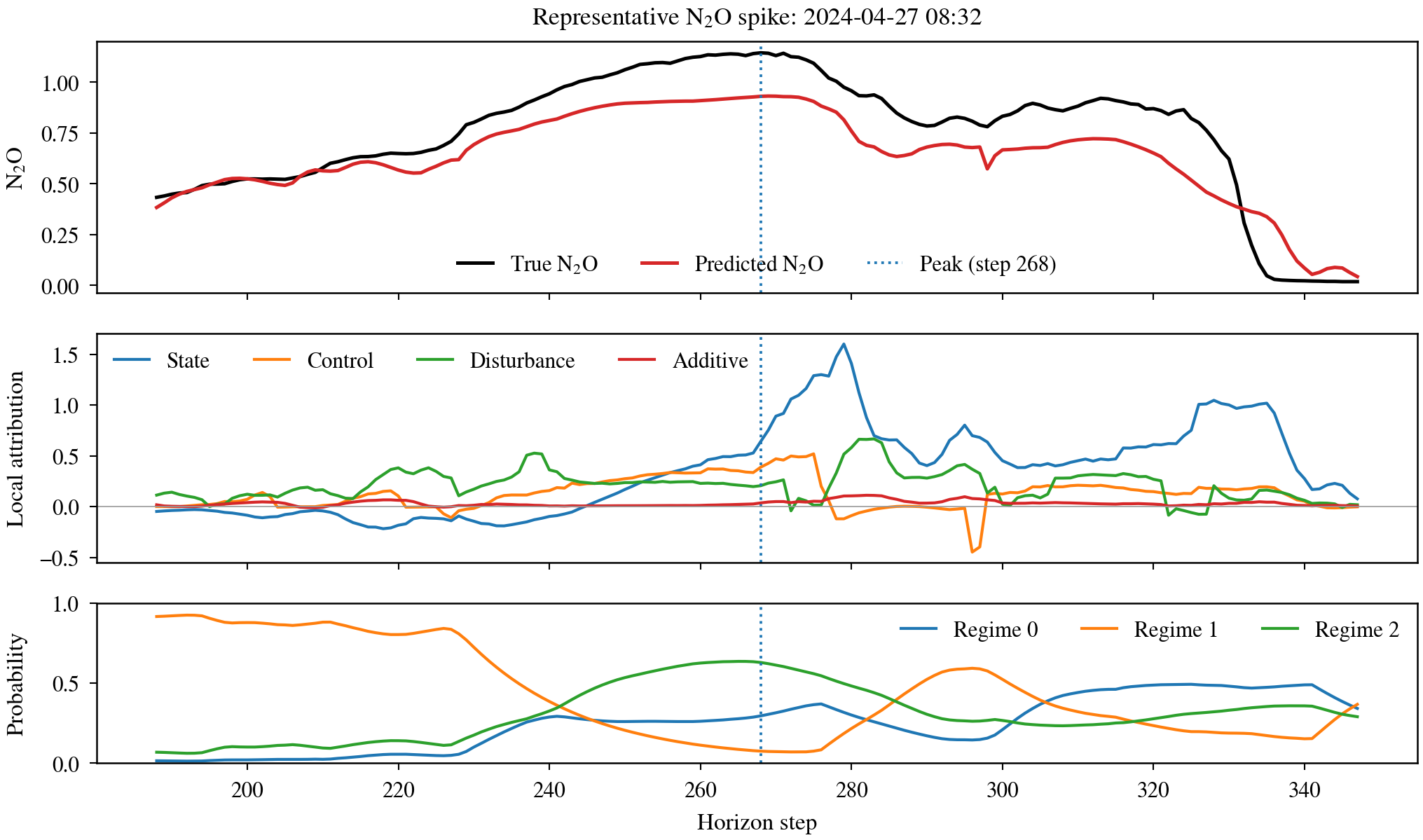}
\caption{Representative N\textsubscript{2}O spike: 2024-04-27 08:32,
peak at horizon step 268, regime~2 dominant.  Per-channel
\emph{magnitude} contribution at the peak (sum of absolute driver
contributions within each channel, in raw $\Delta z$ units, not
normalised to a share over channels): state $0.64$, control $0.39$,
disturbance $0.21$, additive $0.04$, residual $0.00$.  Named top
drivers: N\textsubscript{2}O (state, persistence), DO setpoint
(control, low-DO formation pathway), process temperature (disturbance).
These are model-relative attributions, not proof of physical causality.}
\label{fig:spike-case}
\end{figure*}

\noindent\textbf{N\textsubscript{2}O spike attribution (Aved{\o}re).}
N\textsubscript{2}O is a potent greenhouse gas
($\approx 265 \times$ CO\textsubscript{2} warming
potential~\citep{ravishankara2009n2o}) emitted during biological
wastewater treatment through coupled nitrification/denitrification
pathways~\citep{kampschreur2009n2o}.  Returning to the Aved{\o}re
benchmark, we analyse 24 separated N\textsubscript{2}O spike events
using a reproducible local-maximum, prominence, and
minimum-separation selection rule on the evaluation set
(Fig.~S2 in the supplementary material).  Attribution is stable across events: at spike
apex, the dominant channel is state in 13/24 events, control in 6/24,
and disturbance in 5/24.  Mean apex shares (each channel's magnitude
normalised by the across-channel magnitude total at the peak) are
state 0.399, control 0.246, disturbance 0.273, additive 0.082, and
residual 0.000.  The
recurring top drivers include N\textsubscript{2}O and NO\textsubscript{3}
on the state side, O\textsubscript{2} setpoint and Valve~1 on the control
side, and influent flow and blower airflow on the disturbance side.
Regime~1 dominates 23/24 apex events.

These attributions are model-relative hypotheses, not proof of physical
causality.  Their value is that they are stable, consistent, and
mechanistically plausible, providing the kind of pattern a process
engineer can engage with rather than a single scalar prediction.

\noindent\textbf{Representative single-spike attribution.}
Fig.~\ref{fig:spike-case} shows the 2024-04-27 08:32 event (peak at
step~268, regime-2 dominant, true peak $1.145$, predicted $0.931$,
window RMSE $0.269$).  This event was selected separately for
case-study representativeness and is not one of the 24 multi-spike
attribution events analysed above; among those 24, the only
non-regime-1 apex is a regime-0 event (2023-12-01).  Named
top-driver-per-channel signed contributions at the peak (in raw
$\Delta z$ units, before normalisation to shares): top state
N\textsubscript{2}O itself ($0.509$, autoregressive persistence);
top control tank-1 dissolved-oxygen set-point ($0.263$, consistent
with low DO promoting N$_2$O formation~\citep{kampschreur2009n2o});
top disturbance process water temperature ($0.236$).  A typical operations dashboard would have
shown ``N$_2$O peak near 1\,mg/L at 08:32''; CCSS-IX produces, for the
same moment, a quantitative, named decomposition of what was driving
the peak.

\noindent\textbf{Causal Isolation Index.}
The structured update law's regime-weighted causal row
$W_{\text{eff}}(t)[j] = \sum_k p_{\text{gate}}(t,k)\,W_k[k,\,\mathrm{N_2O},\,j]$
aggregates each regime's coupling from variable $j$ into
N\textsubscript{2}O, and yields a one-dimensional spike-precursor
signal
$\mathrm{CII}(t) = \sum_{j \ne \mathrm{N_2O}} W_{\text{eff}}(t)[j]\,
|\Delta\hat{y}_j(t)| / w_{\text{norm}}(t)$
with normaliser
$w_{\text{norm}}(t) = \sum_{j \ne \mathrm{N_2O}} W_{\text{eff}}(t)[j]$:
the magnitude of upstream non-N\textsubscript{2}O state changes
weighted by their model-attributed coupling into N\textsubscript{2}O.
The trajectory and the resulting CII signal are uniformly smoothed
with a 5-step window before differencing, and the coupled/isolated
classifier flags a spike as coupled when the z-scored CII exceeds
$1\sigma$ at any point in the 60-step lookback before the spike
onset.  A black-box simulator cannot produce this signal because
$W_{\text{eff}}$ is not exposed; CCSS-IX produces it as a side-effect
of the channel decomposition (Fig.~S8).  On 30 evaluation windows
containing 124 N\textsubscript{2}O spike events, CII labels $112/124$
($90.3\%$) as \emph{coupled} (some non-zero precursor in the lead
window before the peak) and $12/124$ ($9.7\%$) as \emph{isolated}
(autonomous, N\textsubscript{2}O leads); mean lead time on coupled
spikes is $47$ steps ($\approx 94$ min at 2-min sampling).

The cross-lag correlation between CII and N\textsubscript{2}O reaches
its maximum at $r=0.146$ (lag $=24$ steps), a modest predictive
signal.  The coupling label is therefore a precursor-visibility flag
rather than a calibrated predictor; any
operational use would require an explicit threshold and a
precision/recall sweep, which we leave to future closed-loop work.
The same per-channel attribution pattern is preserved when the 24
spike events are stratified by magnitude (Fig.~S2), confirming
robustness across spike severity.


\subsection{Multi-site validation: matched-budget head-to-head on Agtrup}
\label{sec:res-agtrup}

The Aved{\o}re primary benchmark (\S\ref{sec:res-arch}) excluded
recurrent and state-space baselines on principled grounds: irregular
sampling and $42.6\%$ missingness in the locked test slice both make
LSTM and S5 inapplicable without protocol-breaking adaptations (see
the CCSS-RS preprint~\citep{simethy2026ccssrs} for the
baseline-fairness analysis).  The Agtrup dataset
(regular 2-min grid, $0\%$ missingness, two-state nutrient-removal
record over $525{,}600$ samples) satisfies both conditions natively,
so it is the appropriate venue for the like-for-like comparison.  All four
architectures train on the same $70\%/30\%$ split at matched parameter
budget (CCSS-RS / CCSS-IX: $6.9$\,M; LSTM: $1.21$\,M; S5: $1.13$\,M
diagonal-real-valued reference at $d_{\text{hid}}=384$, $d_{\text{state}}=64$,
$n_{\text{layers}}=2$) and are evaluated under the same
10-random-window protocol used for the Aved{\o}re Table~\ref{tab:archladder}.

Across four seeds $\{1, 7, 19, 42\}$ (Table~\ref{tab:agtrup}), CCSS-RS
and CCSS-IX cluster tightly: CCSS-RS retains a narrow NH$_4^+$ RMSE
edge ($0.593\pm0.118$ vs.\ $0.608\pm0.100$) and matches CCSS-IX on
NH$_4^+$ $R^2$ and CRPS inside within-seed noise (gaps $0.004$ and
$0.001$ at SDs of $0.04$--$0.06$); CCSS-IX edges CCSS-RS on PO$_4^{3-}$
RMSE ($0.320 \pm 0.031$ vs.\ $0.339 \pm 0.015$), PO$_4^{3-}$ $R^2$
($0.700$ vs.\ $0.650$, the only gap outside one within-seed SD on
this target), and NLL for both targets ($-0.189$ vs.\ $+0.131$ on
NH$_4^+$, $-0.406$ vs.\ $-0.392$ on PO$_4^{3-}$, both small relative
to their SDs).
LSTM falls substantially behind on every metric: RMSE is $\sim$2$\times$
worse on NH$_4^+$ ($1.174 \pm 0.155$) and the Gaussian likelihood
collapses on test windows, yielding NLL of $+78$ to $+79$.  S5 is
worse still, with RMSE 2--2.5$\times$ above CCSS and $R^2$ negative on
both variables (indicating the diagonal-S5 head-mean is worse than a
constant baseline on this multi-output target).  Together these results
confirm that the architecture-class ordering CCSS $>$ LSTM $>$ S5 is
robust to the recurrent-vs-SSM choice at matched compute.

\noindent\textbf{Structural stability.}
A second observation reinforces the case for interpretable structure.
CCSS-IX completed all $10$ evaluation windows on every seed.  CCSS-RS
at seed~7 diverged on $5$ of $10$ windows, producing non-finite mean
forecasts that required a per-window NaN guard to skip the affected
windows and log them transparently.  The likely cause is mechanical: CCSS-IX's structured fast expert
(the per-regime inner-loop update module, in CCSS-RS terminology)
bounds the per-step increments more conservatively than CCSS-RS's
unconstrained neural residual, and that bound carries over to
inference-time robustness on the adversarial test contexts that drive
divergence.  We retain the partial-window
seed-7 CCSS-RS aggregate in Table~\ref{tab:agtrup} with a footnote
disclosure; because the skipped windows are the harder ones, the
reported CCSS-RS metrics \emph{understate} rather than overstate the
gap to CCSS-IX.

\begin{table}[!htb]
  \centering
  \caption{\textbf{Multi-site head-to-head on Agtrup.}\ Mean~$\pm$~SD
  across four seeds, each evaluated on 10 random test windows. Lower is
  better for MAE, RMSE, CRPS, NLL; higher is better for $R^{2}$.
  \textbf{Bold} marks the best value per row.}
  \label{tab:agtrup}
  \newcommand{\vsd}[2]{\ensuremath{#1 \pm #2}}
  \newcommand{\bsd}[2]{\ensuremath{\mathbf{#1} \pm \mathbf{#2}}}
  \resizebox{\columnwidth}{!}{%
  \begin{tabular}{@{}lcccc@{}}
    \toprule
    \textbf{Metric} & \textbf{CCSS-RS}\textsuperscript{\textdagger}
                    & \textbf{CCSS-IX} & \textbf{LSTM} & \textbf{S5} \\
    \midrule
    \addlinespace[2pt]
    \multicolumn{5}{@{}l}{\textit{NH$_4^+$ (Tank~1)}}\\[1pt]
    MAE     & \vsd{0.380}{0.079}  & \bsd{0.371}{0.062}  & \vsd{0.741}{0.100}  & \vsd{1.001}{0.021}  \\
    RMSE    & \bsd{0.593}{0.118}  & \vsd{0.608}{0.100}  & \vsd{1.174}{0.155}  & \vsd{1.443}{0.086}  \\
    $R^{2}$ & \vsd{0.746}{0.044}  & \bsd{0.750}{0.058}  & \vsd{0.067}{0.173}  & \vsd{-0.418}{0.229} \\
    CRPS    & \vsd{0.294}{0.059}  & \bsd{0.293}{0.052}  & \vsd{0.705}{0.103}  & \vsd{0.974}{0.016}  \\
    NLL     & \vsd{+0.13}{0.15}   & \bsd{-0.19}{0.32}   & \vsd{+78.5}{24.1}   & \vsd{+117}{48}      \\
    \midrule
    \addlinespace[2pt]
    \multicolumn{5}{@{}l}{\textit{PO$_4^{3-}$ (Tank~1)}}\\[1pt]
    MAE     & \vsd{0.192}{0.011}  & \bsd{0.185}{0.013}  & \vsd{0.317}{0.032}  & \vsd{0.444}{0.068}  \\
    RMSE    & \vsd{0.339}{0.015}  & \bsd{0.320}{0.031}  & \vsd{0.553}{0.062}  & \vsd{0.713}{0.065}  \\
    $R^{2}$ & \vsd{0.650}{0.044}  & \bsd{0.700}{0.057}  & \vsd{0.102}{0.206}  & \vsd{-0.475}{0.168} \\
    CRPS    & \vsd{0.150}{0.012}  & \bsd{0.148}{0.017}  & \vsd{0.301}{0.036}  & \vsd{0.434}{0.072}  \\
    NLL     & \vsd{-0.39}{0.66}   & \bsd{-0.41}{0.78}   & \vsd{+79.0}{36.9}   & \vsd{+215}{119}     \\
    \bottomrule
  \end{tabular}}
  \par\vspace{4pt}
  \begin{minipage}{\columnwidth}
    \scriptsize
    \textsuperscript{\textdagger}Seed~7 CCSS-RS aggregates 5 of 10 windows;
    the remaining 5 produced non-finite mean forecasts at inference (Picard
    rollout divergence) and were excluded by a per-window NaN guard. Other
    models and seeds use all 10 windows. The skipped windows are the harder
    subset, so reported CCSS-RS metrics are optimistically biased.
  \end{minipage}
\end{table}

\subsection{Real-plant validity layer: shadow-mode replay on two WWTPs}
\label{sec:res-wwtp}

The two real-plant datasets are most useful because they expose
\emph{opposite} failure modes of support-only abstention
(Table~\ref{tab:shadow}).

\noindent\textbf{Aved{\o}re (unsafe-supported blocking).}
In the locked chronological replay, Aved{\o}re is the unsafe-supported
case.  Support-only selection is too permissive: at unsafe-cost 4,
shadow-mode regret is 68 and unsafe acceptance rate is 0.266.  Strict
certification lowers regret to 39 and unsafe acceptance to 0.047.
Calibrated reopening performs identically to strict on this slice (regret
39), which is the desired behavior because there are no safe unsupported
opportunities to recover at Aved{\o}re.

\noindent\textbf{Agtrup (safe-unsupported recovery).}
Agtrup gives the complementary case.  Support-only selection has
shadow-mode regret 10 and unsafe acceptance rate 0.083.  Strict
certification removes unsafe acceptance but also discards safe
unsupported opportunities (regret 7).  Calibrated reopening keeps unsafe
acceptance at 0.0 while increasing acceptance from 0.542 to 0.625 and
reducing regret to 5.

\noindent\textbf{Why both modes matter.}
Together, these cases demonstrate why support-only abstention is
incomplete.  On Aved{\o}re, support-only is too permissive.  On Agtrup,
strict rejection is safer but too blunt.  A useful WWTP validity layer
must handle both modes.  Table~\ref{tab:shadow} reports
per-plant regret alongside the two-plant aggregate at three unsafe-cost
levels: the 43.6\%-headline aggregate reduction at unsafe-cost~4 (78~$\to$~44) is
\emph{Agtrup-driven}, since calibrated reopen ties strict on Aved{\o}re
(no safe-unsupported recoverable scenarios) while reducing regret from 7
to 5 on Agtrup.  We surface this explicitly to avoid overstating the
aggregate effect.  Fig.~\ref{fig:risk-coverage} renders the same
trade-off as a risk--coverage curve: calibrated reopen dominates
support-only abstention at moderate-to-high unsafe-cost weights, with
strict certification sitting between the two.
\raggedbottom
\begin{table}[!h]
\centering
\footnotesize
\caption{Per-plant and aggregate two-plant shadow-mode regret at three
unsafe-cost weights. Calibrated reopening is the best policy at
unsafe-cost 4 and 8; the aggregate gain is driven by Agtrup. Lower is
better.}
\label{tab:shadow}
\setlength{\tabcolsep}{5pt}
\renewcommand{\arraystretch}{1.15}
\begin{tabular}{lcccc}
\toprule
Plant & \makecell{Unsafe-cost \\ weight} & \makecell{Support-only \\ abstention} & \makecell{Strict \\ certification} & \makecell{Calibrated \\ reopening} \\
\midrule
\multirow{3}{*}{Aved{\o}re}
  & 2 & 34          & \textbf{33} & \textbf{33} \\
  & 4 & 68          & \textbf{39} & \textbf{39} \\
  & 8 & 136         & \textbf{51} & \textbf{51} \\
\midrule
\multirow{3}{*}{Agtrup}
  & 2 & 6  & 7 & \textbf{5} \\
  & 4 & 10 & 7 & \textbf{5} \\
  & 8 & 18 & 7 & \textbf{5} \\
\midrule
\multirow{3}{*}{Aggregate}
  & 2 & 40  & 40 & \textbf{38} \\
  & 4 & 78  & 46 & \textbf{44} \\
  & 8 & 154 & 58 & \textbf{56} \\
\bottomrule
\end{tabular}
\end{table}

\begin{figure*}[!b]
\centering
\includegraphics[width=1\linewidth]{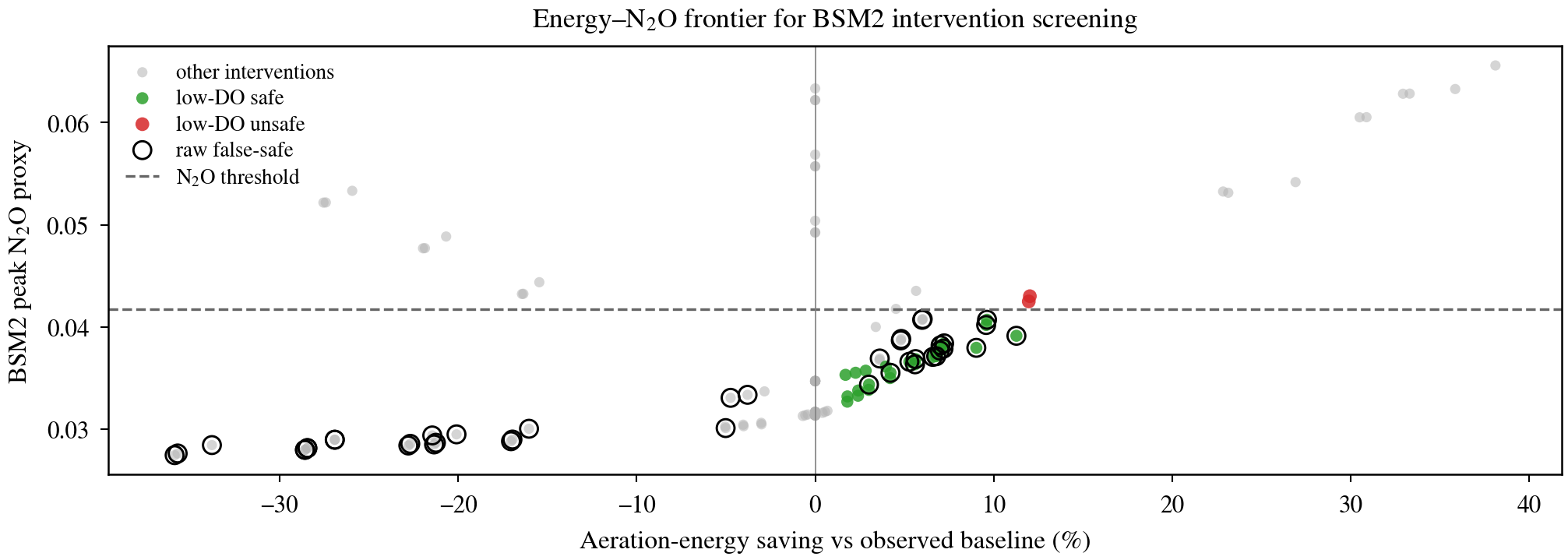}
\caption{Energy--N\textsubscript{2}O frontier for 27 low-DO BSM2
scenarios.  Safe low-DO cases (green) cluster at median aeration-energy
saving 2.41\%.  Unsafe low-DO cases (red) extend to 7.02\% median saving.
Filled circles are raw false-safe approvals: the learned rollout marks
them safe but the BSM2 oracle marks them unsafe.  The validity layer
screens the most attractive energy-saving region, not merely unusual
scenarios.}
\label{fig:energy-n2o}
\end{figure*}

\subsection{BSM2 mechanistic oracle: confirmed-safe screening}
\label{sec:res-bsm2}

Table~\ref{tab:bsm2} reports the scaled BSM2 benchmark on the main
horizon-12 validated slice ($n=3$ scenario groups).  Support-only
selection has chosen-action unsafe rate 1.0 (95\% CP $[0.292,\,1.000]$):
every action it selects is oracle-unsafe.  This is substantive, not a
pipeline artefact: support-only chooses the action closest to the
training distribution, and on this slice the most-supported actions are
the aggressive low-DO interventions that BSM2 marks unsafe.  Strict
certification and calibrated reopening both reduce chosen-action unsafe
rate to 0.0; calibrated reopening further lowers mean regret from 2.984
to 2.912 by recovering utility without reopening unsafe choices.  We
caution that with $n=3$ groups the 1.0 support-only rate is
exact-on-this-slice rather than a robust population estimate; the
symmetric caveat is that the strict/calibrated 0.0 rate has the same
small-$n$ CI ($[0.000,0.708]$ at 95\% confidence), so the slice alone
only excludes true unsafe rates above $\sim 70\%$.  The qualitative
pattern (support filtering anti-correlated with safety) is corroborated
by the 27-scenario frontier in Fig.~\ref{fig:energy-n2o} and the
$n=187$ witness matrix in Table~\ref{tab:witnesses}, on which the
quantitative claims rest.  At horizon~16 the strict and calibrated
policies still retain unsafe rate $0.333$, which we report as the
validated boundary of the witness regime rather than smoothing it
away.

\begin{table}[!htb]
\centering
\footnotesize
\caption{BSM2 counterfactual benchmark (horizon-12 main validated slice,
$n=3$ scenario groups).  Strict and calibrated reopening eliminate
unsafe chosen actions; calibrated reopening further lowers regret.
Lower is better.  95\% Clopper--Pearson interval in brackets.}
\label{tab:bsm2}
\setlength{\tabcolsep}{3pt}
\begin{tabular}{lccc}
\toprule
Policy & Unsafe rate (95\% CP) & Acted & Regret \\
\midrule
Unfiltered (best pred.)     & 0.333 $[0.008,\,0.906]$ & 1.000 & 1.983 \\
Support-only                & 1.000 $[0.292,\,1.000]$ & 1.000 & 2.661 \\
Strict certification        & 0.000 $[0.000,\,0.708]$ & 1.000 & 2.984 \\
\textbf{Calibrated reopen}  & \textbf{0.000 $[0.000,\,0.708]$} & \textbf{1.000} & \textbf{2.912} \\
Always-abstain              & ---                     & 0.000 & 3.201 \\
\bottomrule
\end{tabular}
\end{table}

\begin{figure*}[!t]
\centering
\includegraphics[width=1\linewidth]{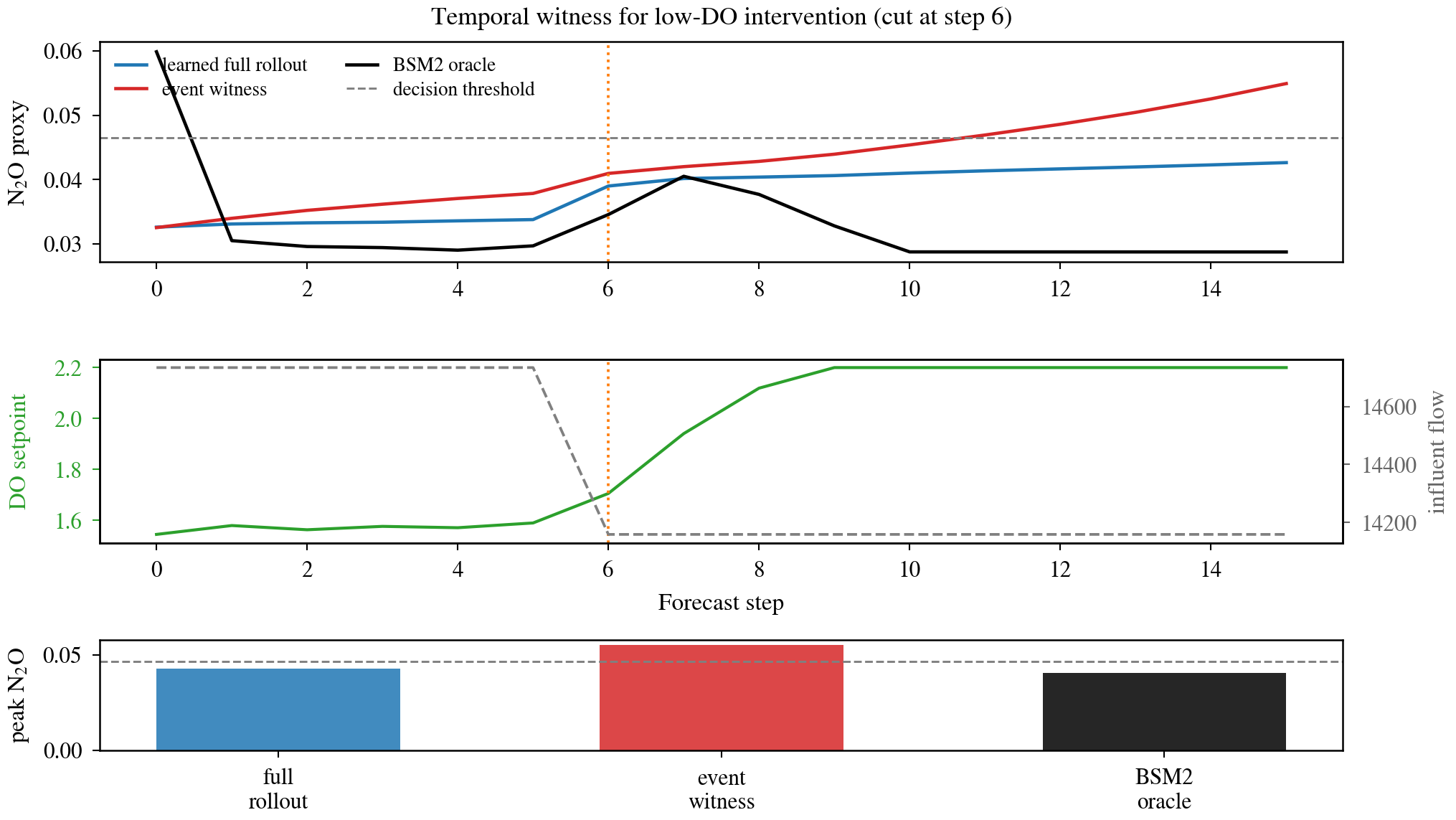}
\caption{Representative temporal witness for a single low-DO
intervention scenario (BSM2 scenario library identifier
low-DO strategy~04).  The raw full rollout (dashed blue,
peak 0.0426) remains below the N\textsubscript{2}O threshold (grey).
The event-aligned witness partition $[0$-$6$-$16]$ (solid orange, peak
0.0549) exceeds the threshold.  The BSM2 oracle (solid green, peak
0.0599) confirms the unsafe outcome.  The disagreement between full
rollout and witness is attributed to the aeration-change event at
step~6, visible in CCSS-IX's control influence channel.}
\label{fig:witness}
\end{figure*}

\subsection{Energy--N\textsubscript{2}O frontier and temporal witnesses}
\label{sec:res-witness}

\noindent\textbf{Energy--N\textsubscript{2}O frontier.}
Fig.~\ref{fig:energy-n2o} shows 27 low-DO BSM2 scenarios.  Safe low-DO
scenarios have median aeration-energy saving of 2.41\%; unsafe ones
extend to 7.02\%.  Of 17 oracle-unsafe low-DO scenarios, 15
are \emph{raw false-safe}: the learned rollout marks them safe, but
the BSM2 oracle exceeds the N\textsubscript{2}O threshold.  The
validity layer thus screens a real operational tension: the
\emph{attractive} energy-saving region concentrates unsafe outcomes,
consistent with the greenhouse-gas-aware control
literature~\citep{flores2011ghg,wst2025n2ogann}.

\noindent\textbf{Temporal witnesses.}
Table~\ref{tab:witnesses} summarises the 16-seed h16 low-DO witness
matrix.  Raw-full hard gating prevents 0/187 false-safe approvals
(the full rollout cannot self-falsify by definition); dyadic
decomposition prevents 20/187 and matched random B4 prevents 19/187.
Event-aligned B4 prevents 93/187 (95\% CP $[42.4\%,\,57.1\%]$),
$4.65\times$ the dyadic baseline, while false-blocking only 16/653
true-safe raw approvals.  Decision-stressed B4 is an active-search
upper bound at 130/187.  A worst-case paired McNemar test of
event-aligned versus dyadic on the same 187 scenarios yields
$p = 2 \times 10^{-22}$, so the gap is statistically robust
regardless of the unobserved exact pairing.  The same dominance survives
multiplicity correction at $\alpha = 0.01$: applying Holm--Bonferroni to
pairwise McNemar tests of event-aligned B4 against each of the four
contrast families in Table~\ref{tab:witnesses} (raw-full, dyadic,
matched-random B4, decision-stressed B4) keeps every comparison
significant.

The representative witness (Fig.~\ref{fig:witness}) is a single
low-DO intervention from the BSM2 scenario library (identifier
low-DO strategy~04): raw full
rollout peaks at 0.0426 (below threshold 0.0465), event-aligned
partition $[0$-$6$-$16]$ reaches 0.0549, BSM2 oracle 0.0599.  The
disagreement is attributed via CCSS-IX's control channel to the
aeration-change event at step~6.
The harder h24 horizon is retained as a boundary condition where
event-aligned B4 prevents 0/194 false-safe approvals while
decision-stressed B4 prevents 48/194.  We report this transparently
because it defines the validated regime: h16 event-aligned witnessing
is strong; h24 requires a different event family or active partition
search.

\begin{table}[!htb]
\centering
\footnotesize
\caption{Temporal witness comparison (16-seed h16 low-DO matrix, $n=187$).
Event-aligned B4 prevents $4.65\times$ as many false-safe approvals as
dyadic. Worst-case paired McNemar (event-aligned vs dyadic) yields
$p < 10^{-21}$. CP $=$ 95\% Clopper--Pearson interval.}
\label{tab:witnesses}
\setlength{\tabcolsep}{4pt}
\renewcommand{\arraystretch}{1.15}
\resizebox{\columnwidth}{!}{%
\begin{tabular}{lcc}
\toprule
Witness family & \makecell{False-safe prevented \\ (95\% CP)} & \makecell{True-safe \\ blocked} \\
\midrule
Raw-full hard gate                   & 0 \,[0.0\%, 1.95\%]            & 0  \\
Dyadic decomposition                 & 20 \,[6.7\%, 16.0\%]           & 9  \\
Matched random B4                    & 19 \,[6.2\%, 15.4\%]           & 6  \\
\textbf{Event-aligned B4 (main)}     & \textbf{93 \,[42.4\%, 57.1\%]} & 16 \\
Decision-stressed B4 (upper)         & 130 \,[62.3\%, 76.4\%]         & 20 \\
\bottomrule
\end{tabular}}
\end{table}


\section{Discussion}
\label{sec:discussion}

The structured-update and self-falsifying-validity layers are one
contribution, not two.  The Aved{\o}re N$_2$O attribution illustrates
the synergy: stable state and control channel dominance with zero
residual share across 24 spike events means that when the validity
layer flags a low-DO scenario with an event-aligned witness, the
structured channels attribute the disagreement to a specific control
influence.  That is the difference between ``scenario refused'' and
``scenario refused because the aeration change at step~6 falsifies
the approval''~\citep{rudin2019stop}.  The
energy--N$_2$O frontier result follows for the same reason: an opaque
uncertainty score offers no narrative for why a near-optimal
energy-saving intervention is unsafe; a channel-attributed witness
does.

The choice of main model follows from this same logic.  All three
CCSS-IX variants are statistically indistinguishable from CCSS-RS on
RMSE (Table~\ref{tab:archladder}); we adopt the adaptive
context-conditioned variant not on fidelity grounds but because the
downstream channel-attribution analyses (\S\ref{sec:res-interp})
operate against its $A_k(\gamma_t), B_k(\gamma_t), E_k(\gamma_t)$
couplings.  The combined architecture also handles both real-plant
failure modes (unsafe-supported approvals at Aved{\o}re and
safe-unsupported rejections at Agtrup) without per-plant tuning,
which is the practical bar for operator-facing deployment.

\subsection{Limitations}
\label{sec:limits}

All real-plant evaluations are \emph{shadow-mode}: the evidence supports
screening and decision support, not closed-loop operational benefit.
The h24 temporal-witness stress test shows that event-aligned
B4 witnessing is not horizon-invariant; the validated regime is h16.
Calibrated reopening depends on the calibration block's
representativeness and on its sample size: at the validity thresholds
used here the calibration block is small enough that the calibration
quantile itself carries non-negligible variance, and distribution
shift, sensor drift, and process changes must be monitored before
operational use~\citep{cleaveland2024robust,shift2026generative,
robust2026staggered,sin2009uncertainty}.  The event-aligned witness
mechanism is also potentially evadable by an adversarial operator who
constructs control trajectories without large normalised changes at
decision-relevant points; defending against this requires either an
active partition search (the decision-stressed B4 row of
Table~\ref{tab:witnesses} provides an upper bound) or a learned
witness-policy on top of the structured channels.

Per-seed dispersion on this benchmark is wide ($-9\%$ to $+17\%$ for
Static, $-6\%$ to $+9\%$ for Adaptive), so any fidelity-headline claim
requires a pre-registered evaluation seed set; we recommend reporting
paired-CI parity rather than ratio-of-means percentages.  The
per-plant decomposition in Table~\ref{tab:shadow} similarly makes it
explicit that the $43.6\%$ aggregate-regret reduction at unsafe-cost~4
is Agtrup-driven; the per-plant table is preferable to leading with
the aggregate.

The like-for-like recurrent and SSM head-to-head is conducted on the
regular-grid Agtrup dataset (\S\ref{sec:res-agtrup},
Table~\ref{tab:agtrup}); LSTM~\citep{hansen2024lstmn2o} and a
diagonal-real-valued S5 reference~\citep{smith2023s5} both fall behind
the CCSS family at matched compute, and the architecture-class
ordering is robust.  Canonical Mamba~\citep{gu2023mamba} is left out
because the sampling-regularity argument from the CCSS-RS
preprint~\citep{simethy2026ccssrs} disqualifies its assumed regular,
fully-observed sampling on Aved{\o}re without protocol-breaking
adaptations; on Agtrup, where regular sampling and complete data both
hold, the diagonal-S5 reference establishes the SSM-class ceiling and
we expect a tuned Mamba to land in the same regime.  Finally, all quantitative evaluation is on wastewater data.  In
exploratory tests on episodic batch processes (IndPenSim fed-batch,
batch distillation), the structured fast expert's Picard rollout
diverged: the inverse log-magnitude transform saturates when targets
span very different scales.  This is a magnitude-scaling pipeline
issue, not a structural-decomposition issue.  Full multi-domain
validation, via a domain-aware re-scaling layer, is future work.

\subsection{Deployment pathway and future work}
\label{sec:future}

The immediate deployment path is operator-in-the-loop shadow screening:
a plant uses the locked protocol to label candidate interventions as
\textsc{accept}, \textsc{abstain}, \textsc{reopen}, or
\textsc{witness} (the four outcomes of Alg.~\ref{alg:validity}) before
any physical control change.  The event-aligned witness is the natural
operator-facing explanation because it frames the outcome in events
the operator already understands (aeration changes, influent pulses,
phase transitions)~\citep{seborg2017process,qin2012survey}.  Future
work spans (i)~a canonical-Mamba comparison on Agtrup once a stable
build is available on the deployment stack (the
diagonal-S5 ceiling established in
Table~\ref{tab:agtrup} bounds the expected lift);
(ii)~a phase-aware variant of the CCSS scaffold for episodic batch
processes (separate paper, non-trivial inductive-bias change);
(iii)~closed-loop deployment with operator opt-in and a streaming-aware
calibration update under distribution shift~\citep{cleaveland2024robust,
shift2026generative,robust2026staggered}.

\FloatBarrier

\section{Conclusion}
\label{sec:conclusion}

Wastewater digital twins used for intervention screening need two
properties most existing approaches provide separately: interpretable
simulation dynamics that expose how controls and disturbances drive
predicted outcomes, and certified intervention decisions that are
stable under temporal re-evaluation.

We introduced CCSS-IX, an adaptive context-conditioned structured
simulator that achieves statistical fidelity parity with the black-box
CCSS-RS reference across ten seeds (paired 95\% bootstrap CI
$[-0.010,+0.024]$, pct $+1.08\%$).  The static and adaptive+residual
variants of the same structured family land in the same parity band
($[-0.029,+0.038]$ and $[-0.010,+0.050]$ respectively).  Across the
architecture ladder, interpretable structured channels exposing direct
state, control, disturbance, and nonlinear-response decompositions
(Table~\ref{tab:archladder}) are therefore obtained at no measurable
fidelity cost relative to the black-box scaffold.

Built on this structured foundation, the self-falsifying validity layer
produces four-outcome decisions that reduce two-plant shadow-mode regret
by 43.6\% relative to support-only abstention (Agtrup-driven;
Table~\ref{tab:shadow}), eliminate unsafe chosen actions on the BSM2
main validated slice (Table~\ref{tab:bsm2}), and prevent 93/187 raw
false-safe N\textsubscript{2}O approvals via event-aligned temporal
witnesses ($4.65\times$ the dyadic baseline; worst-case paired McNemar
$p<10^{-21}$; Table~\ref{tab:witnesses}).  The witness attribution to
specific control and disturbance events is made possible by CCSS-IX's
interpretable structure: support filtering and partition-defect
screening can wrap any open-loop simulator, but channel-level
attribution requires the structured decomposition.

\section*{Data and Code Availability}

The Aved{\o}re N\textsubscript{2}O dataset is available at Mendeley
Data (DOI \texttt{10.17632/xmbxhscgpr.4})~\citep{hansen2024data} and
the Agtrup/BlueKolding dataset at Mendeley Data
(DOI \texttt{10.17632/34rpmsxc4z.1})~\citep{mohammadi2024data}.
Redistribution of local raw plant CSV snapshots is subject to plant and
data-owner permissions.  The CCSS-RS simulator foundation is described
in the public preprint of~\citet{simethy2026ccssrs}.  The CCSS-IX
training and validity-layer source code, together with the BSM2 scenario
outputs underlying Table~\ref{tab:bsm2}, the 16-seed horizon-16 low-DO
witness matrix underlying Table~\ref{tab:witnesses}, shadow-mode
decision logs, the energy--N\textsubscript{2}O frontier, risk-coverage
audit tables, representative temporal-witness data, split indices, and
configuration metadata, will be released at an anonymous mirror upon
acceptance.

\section*{Declaration of Generative AI Use}

During the preparation of this work, Claude (Anthropic) was used for
language editing.  The authors reviewed and edited all content and take
full responsibility for the published article.

\section*{CRediT Author Contribution Statement}

\textbf{Gary Simethy:} Conceptualization, Methodology, Software,
Writing -- original draft.
\textbf{Daniel Ortiz Arroyo:} Supervision, Methodology, Writing --
review and editing.
\textbf{Petar Durdevic:} Supervision, Resources, Writing -- review and
editing.

\section*{Declaration of Competing Interest}

The authors declare no competing financial interests or personal
relationships that could have influenced the work reported in this
paper.

\section*{Funding}

This research was supported by Aalborg University and Helix Lab in
Denmark under the Novo Nordisk Fonden through project grant number
224611.

\bibliographystyle{elsarticle-num-names}
\bibliography{references}

@article{wang2024dt,
  author    = {Wang, An-Jie and Li, He and He, Zhong and Tao, Yi and
               Wang, Hao and Yang, Min and Savic, Dragan and Daigger, Glen T.
               and Ren, Nanqi},
  title     = {Digital Twins for Wastewater Treatment: A Technical Review},
  journal   = {Engineering},
  volume    = {36},
  pages     = {21--35},
  year      = {2024},
  doi       = {10.1016/j.eng.2024.04.012}
}

@article{rasheed2020dt,
  author    = {Rasheed, Adil and San, Omer and Kvamsdal, Trond},
  title     = {Digital Twin: Values, Challenges and Enablers from a Modeling
               Perspective},
  journal   = {IEEE Access},
  volume    = {8},
  pages     = {21980--22012},
  year      = {2020},
  doi       = {10.1109/ACCESS.2020.2970143}
}

@article{tao2019dt,
  author    = {Tao, Fei and Zhang, Meng and Liu, Yushan and Nee, Andrew Y. C.},
  title     = {Digital twin driven prognostics and health management for complex
               equipment},
  journal   = {CIRP Annals},
  volume    = {67},
  number    = {1},
  pages     = {169--172},
  year      = {2018},
  doi       = {10.1016/j.cirp.2018.04.055}
}

@article{mehta2025watersector,
  author    = {Ghorbani Bam, Pooria and Rezaei, Nader and Roubanis, Alexander
               and Austin, Dana and Austin, Elinor and Tarroja, Brian
               and Takacs, Imre and Villez, Kris and Rosso, Diego},
  title     = {Digital Twin Applications in the Water Sector: A Review},
  journal   = {Water (MDPI)},
  volume    = {17},
  number    = {20},
  pages     = {2957},
  year      = {2025},
  doi       = {10.3390/w17202957}
}

@article{simethy2026ccssrs,
  author    = {Simethy, Gary and Ortiz Arroyo, Daniel and Durdevic, Petar},
  title     = {Data-Driven Open-Loop Simulation for Digital-Twin Operator
               Decision Support in Wastewater Treatment},
  journal   = {arXiv preprint arXiv:2604.20935},
  year      = {2026},
  doi       = {10.48550/arXiv.2604.20935}
}

@article{hansen2024data,
  author    = {Hansen, Lasse Dalbro and Rani, Archana and
               Stokholm-Bjerregaard, Mikkel A. and Stentoft, Peter A. and
               Ortiz-Arroyo, Daniel and Durdevic, Petar},
  title     = {Time Series Dataset for Modeling and Forecasting of
               {N\textsubscript{2}O} in Wastewater Treatment},
  journal   = {arXiv preprint arXiv:2407.05959},
  year      = {2024},
  doi       = {10.48550/arXiv.2407.05959}
}

@article{mohammadi2024data,
  author    = {Mohammadi, Ehsan and Rani, Archana and Stokholm-Bjerregaard,
               Mikkel and Ortiz Arroyo, Daniel and Durdevic, Petar},
  title     = {Wastewater Treatment Plant Data for Nutrient Removal System},
  journal   = {arXiv preprint arXiv:2407.05346},
  year      = {2024},
  doi       = {10.48550/arXiv.2407.05346}
}

@article{newhart2019review,
  author    = {Newhart, Kyle B. and Holloway, Ryan W. and Hering, Amanda S.
               and Cath, Tzahi Y.},
  title     = {Data-driven performance analyses of wastewater treatment plants:
               A review},
  journal   = {Water Research},
  volume    = {157},
  pages     = {498--513},
  year      = {2019},
  doi       = {10.1016/j.watres.2019.03.030}
}

@article{jeppsson2007bsm2,
  author    = {Jeppsson, Ulf and Pons, Marie-No{\"e}lle and Nopens, Ingmar and
               Alex, Jens and Copp, John B. and Gernaey, Krist V. and
               Rosen, Christian and Steyer, Jean-Philippe and
               Vanrolleghem, Peter A.},
  title     = {Benchmark simulation model no.~2: general protocol and
               exploratory case studies},
  journal   = {Water Science and Technology},
  volume    = {56},
  number    = {8},
  pages     = {67--78},
  year      = {2007},
  doi       = {10.2166/wst.2007.604}
}

@techreport{henze2000asm,
  author      = {Henze, Mogens and Gujer, Willi and Mino, Takashi and
                 van Loosdrecht, Mark C. M.},
  title       = {Activated Sludge Models ASM1, ASM2, ASM2d and ASM3},
  institution = {IWA Publishing},
  type        = {IWA Scientific and Technical Report},
  number      = {9},
  year        = {2000}
}

@article{sin2009uncertainty,
  author    = {Sin, Gurkan and Gernaey, Krist V. and Neumann, Marc B. and
               van Loosdrecht, Mark C. M. and Gujer, Willi},
  title     = {Uncertainty analysis in {WWTP} model applications:
               A critical discussion using an example from design},
  journal   = {Water Research},
  volume    = {43},
  number    = {11},
  pages     = {2894--2906},
  year      = {2009},
  doi       = {10.1016/j.watres.2009.03.048}
}

@article{kampschreur2009n2o,
  author    = {Kampschreur, Marlies J. and Temmink, Hardy and Kleerebezem,
               Robbert and Jetten, Mike S. M. and van Loosdrecht, Mark C. M.},
  title     = {Nitrous oxide emission during wastewater treatment},
  journal   = {Water Research},
  volume    = {43},
  number    = {17},
  pages     = {4093--4103},
  year      = {2009},
  doi       = {10.1016/j.watres.2009.03.001}
}

@article{ravishankara2009n2o,
  author    = {Ravishankara, A. R. and Daniel, John S. and Portmann, Robert W.},
  title     = {Nitrous Oxide ({N\textsubscript{2}O}): The Dominant
               Ozone-Depleting Substance Emitted in the 21st Century},
  journal   = {Science},
  volume    = {326},
  number    = {5949},
  pages     = {123--125},
  year      = {2009},
  doi       = {10.1126/science.1176985}
}

@article{flores2011ghg,
  author    = {Flores-Alsina, Xavier and Corominas, Lluis and Snip, Laura and
               Vanrolleghem, Peter A.},
  title     = {Including greenhouse gas emissions during benchmarking of
               wastewater treatment plant control strategies},
  journal   = {Water Research},
  volume    = {45},
  number    = {16},
  pages     = {4700--4710},
  year      = {2011},
  doi       = {10.1016/j.watres.2011.04.040}
}

@article{hansen2024lstmn2o,
  author    = {Seshan, Siddharth and Poinapen, Johann and Zandvoort, Marcel H. and
               van Lier, Jules B. and Kapelan, Zoran},
  title     = {Forecasting nitrous oxide emissions from a full-scale
               wastewater treatment plant using {LSTM}-based deep learning
               models},
  journal   = {Water Research},
  volume    = {268},
  pages     = {122754},
  year      = {2025},
  doi       = {10.1016/j.watres.2024.122754}
}

@article{mdpi2025symmetryn2o,
  author    = {Huang, Zhengze and Bai, Yuqi and Liu, Hengyu},
  title     = {Symmetry-Inspired Prediction of Nitrous Oxide Emissions in
               Wastewater Treatment Using Deep Learning and Explainable
               Analysis},
  journal   = {Symmetry (MDPI)},
  volume    = {17},
  number    = {2},
  pages     = {297},
  year      = {2025},
  doi       = {10.3390/sym17020297}
}

@article{wst2025n2ogann,
  author    = {Freyschmidt, Arne and K{\"o}ster, Stephan},
  title     = {Novel approach for {AI}-based {N$_2$O} emission reduction in
               biological wastewater treatment relying on genetic algorithms
               and neural networks},
  journal   = {Water Science and Technology},
  volume    = {91},
  number    = {10},
  pages     = {1172--1184},
  year      = {2025},
  doi       = {10.2166/wst.2025.060}
}

@inproceedings{chen2018neuralode,
  author    = {Chen, Ricky T. Q. and Rubanova, Yulia and Bettencourt, Jesse
               and Duvenaud, David K.},
  title     = {Neural Ordinary Differential Equations},
  booktitle = {Advances in Neural Information Processing Systems},
  volume    = {31},
  year      = {2018}
}

@inproceedings{kidger2020ncde,
  author    = {Kidger, Patrick and Morrill, James and Foster, James and
               Lyons, Terry},
  title     = {Neural Controlled Differential Equations for Irregular Time
               Series},
  booktitle = {Advances in Neural Information Processing Systems},
  volume    = {33},
  year      = {2020}
}

@inproceedings{lim2021tft,
  author    = {Lim, Bryan and Ar{\i}k, Sercan {\"O}. and Loeff, Nicolas and
               Pfister, Tomas},
  title     = {Temporal Fusion Transformers for Interpretable Multi-horizon
               Time Series Forecasting},
  journal   = {International Journal of Forecasting},
  volume    = {37},
  number    = {4},
  pages     = {1748--1764},
  year      = {2021},
  doi       = {10.1016/j.ijforecast.2021.03.012}
}

@inproceedings{gu2022s4,
  author    = {Gu, Albert and Goel, Karan and R{\'e}, Christopher},
  title     = {Efficiently Modeling Long Sequences with Structured State Spaces},
  booktitle = {International Conference on Learning Representations},
  year      = {2022}
}

@article{rudin2019stop,
  author    = {Rudin, Cynthia},
  title     = {Stop explaining black box machine learning models for high stakes
               decisions and use interpretable models instead},
  journal   = {Nature Machine Intelligence},
  volume    = {1},
  number    = {5},
  pages     = {206--215},
  year      = {2019},
  doi       = {10.1038/s42256-019-0048-x}
}

@inproceedings{agarwal2021nam,
  author    = {Agarwal, Rishabh and Melnick, Levi and Frosst, Nicholas and
               Zhang, Xuezhou and Lengerich, Ben and Caruana, Rich and
               Hinton, Geoffrey E.},
  title     = {Neural Additive Models: Interpretable Machine Learning with
               Neural Nets},
  booktitle = {Advances in Neural Information Processing Systems},
  volume    = {34},
  year      = {2021}
}

@inproceedings{lundberg2017shap,
  author    = {Lundberg, Scott M. and Lee, Su-In},
  title     = {A Unified Approach to Interpreting Model Predictions},
  booktitle = {Advances in Neural Information Processing Systems},
  volume    = {30},
  year      = {2017}
}

@inproceedings{linderman2017rslds,
  author    = {Linderman, Scott W. and Johnson, Matthew J. and Miller, Andrew C.
               and Adams, Ryan P. and Blei, David M. and Paninski, Liam},
  title     = {Bayesian Learning and Inference in Recurrent Switching Linear
               Dynamical Systems},
  booktitle = {Artificial Intelligence and Statistics},
  pages     = {914--922},
  year      = {2017}
}

@article{fox2011sticky,
  author    = {Fox, Emily B. and Sudderth, Erik B. and Jordan, Michael I. and
               Willsky, Alan S.},
  title     = {A sticky {HDP-HMM} with application to speaker diarization},
  journal   = {Annals of Applied Statistics},
  volume    = {5},
  number    = {2A},
  pages     = {1020--1056},
  year      = {2011},
  doi       = {10.1214/10-AOAS395}
}

@article{balsellsrodas2026rslds,
  author    = {Balsells-Rodas, Carles and Xiang, Zhengrui and Sumba, Xavier
               and Li, Yingzhen},
  title     = {End-to-End Identifiable and Consistent Recurrent Switching
               Dynamical Systems},
  journal   = {arXiv preprint arXiv:2605.06315},
  year      = {2026},
  doi       = {10.48550/arXiv.2605.06315}
}

@article{zhang2025identifiable,
  author    = {Zhang, Congxi and Xie, Yongchun},
  title     = {Identifiable Representation and Model Learning for Latent
               Dynamic Systems},
  journal   = {arXiv preprint arXiv:2410.17882},
  year      = {2024},
  doi       = {10.48550/arXiv.2410.17882}
}

@article{sertbas2025lpvssm,
  author    = {Sertba{\c{s}}, Ahmet Eren and Kumbasar, Tufan},
  title     = {Stable-by-Design Neural Network-Based {LPV} State-Space Models
               for System Identification},
  journal   = {arXiv preprint arXiv:2510.24757},
  year      = {2025},
  doi       = {10.48550/arXiv.2510.24757}
}

@article{mansur2026solis,
  author    = {Mansur, Md.\ Hasanul and Kumbasar, Tufan},
  title     = {{SOLIS}: Physics-Informed Learning of Interpretable Neural
               Surrogates for Nonlinear Systems},
  journal   = {arXiv preprint arXiv:2604.14879},
  year      = {2026},
  doi       = {10.48550/arXiv.2604.14879}
}

@article{zhang2025switchnss,
  author    = {Zhang, Yanxin and Yu, Chengpu and Fabiani, Filippo},
  title     = {Neural Network-Based Identification of State-Space Switching
               Nonlinear Systems},
  journal   = {arXiv preprint arXiv:2503.10114},
  year      = {2025},
  doi       = {10.48550/arXiv.2503.10114}
}

@article{martinez2026doaeration,
  author    = {Martinez De La Hoz, Jeimy L. and Bappy, Mahathir Mohammad
               and Islam, MD Shafikul and Marcantel, Mason and
               Hayes, Michael P.},
  title     = {Interpretable Forecasting of Dissolved Oxygen Leveraging a
               Foundation Model for Proactive Aeration in Rural Wastewater
               Treatment Systems},
  journal   = {Water Research},
  year      = {2026},
  doi       = {10.1016/j.watres.2025.124931}
}

@article{yin2025probwwtp,
  author    = {Yin, Hailong and Chen, Yongqi and Zhou, Jingshu and Xie,
               Yifan and Wei, Qing and Xu, Zuxin},
  title     = {A Probabilistic Deep Learning Approach to Enhance the
               Prediction of Wastewater Treatment Plant Effluent Quality
               under Shocking Load Events},
  journal   = {Water Research X},
  year      = {2025},
  doi       = {10.1016/j.wroa.2024.100291}
}

@article{bohn2024denit,
  author    = {B{\o}hn, Eivind and Eidnes, S{\o}lve and Jonassen, Kjell Rune},
  title     = {Machine Learning in Wastewater Treatment: Insights from
               Modelling a Pilot Denitrification Reactor},
  journal   = {arXiv preprint arXiv:2412.14030},
  year      = {2024},
  doi       = {10.48550/arXiv.2412.14030}
}

@article{lindemann2025lecps,
  author    = {Lin, Vivian and Kaur, Ramneet and Yang, Yahan and
               Dutta, Souradeep and Kantaros, Yiannis and Roy, Anirban and
               Jha, Susmit and Sokolsky, Oleg and Lee, Insup},
  title     = {Safety Monitoring for Learning-Enabled Cyber-Physical
               Systems in Out-of-Distribution Scenarios},
  journal   = {arXiv preprint arXiv:2504.13478},
  year      = {2025},
  doi       = {10.48550/arXiv.2504.13478}
}

@article{xu2025scrc,
  author    = {Xu, Yunpeng and Guo, Wenge and Wei, Zhi},
  title     = {Selective Conformal Risk Control},
  journal   = {arXiv preprint arXiv:2512.12844},
  year      = {2025},
  doi       = {10.48550/arXiv.2512.12844}
}

@article{aponte2023rlwwtp,
  author    = {Aponte-Rengifo, Oscar and Francisco, Mario and Vilanova,
               Ram{\'o}n and Vega, Pastora and Revollar, Silvana},
  title     = {Intelligent Control of Wastewater Treatment Plants Based on
               Model-Free Deep Reinforcement Learning},
  journal   = {Processes},
  volume    = {11},
  number    = {8},
  pages     = {2269},
  year      = {2023},
  doi       = {10.3390/pr11082269}
}

@book{ljung1999sysid,
  author    = {Ljung, Lennart},
  title     = {System Identification: Theory for the User},
  edition   = {2nd},
  publisher = {Prentice Hall},
  address   = {Upper Saddle River, NJ},
  year      = {1999}
}

@book{seborg2017process,
  author    = {Seborg, Dale E. and Edgar, Thomas F. and Mellichamp,
               Duncan A. and Doyle, Francis J.},
  title     = {Process Dynamics and Control},
  edition   = {4th},
  publisher = {Wiley},
  year      = {2017}
}

@inproceedings{geifman2017selective,
  author    = {Geifman, Yonatan and El-Yaniv, Ran},
  title     = {Selective Classification for Deep Neural Networks},
  booktitle = {Advances in Neural Information Processing Systems},
  volume    = {30},
  year      = {2017}
}

@article{angelopoulos2022crc,
  author    = {Angelopoulos, Anastasios N. and Bates, Stephen and Fisch, Adam
               and Lei, Lihua and Schuster, Tal},
  title     = {Conformal Risk Control},
  journal   = {arXiv preprint arXiv:2208.02814},
  year      = {2022},
  doi       = {10.48550/arXiv.2208.02814}
}

@article{robust2026staggered,
  author    = {Cherenson, Daniel M. and Panagou, Dimitra},
  title     = {Staggered Integral Online Conformal Prediction for Safe
               Dynamics Adaptation with Multi-Step Coverage Guarantees},
  journal   = {arXiv preprint arXiv:2604.06058},
  year      = {2026},
  doi       = {10.48550/arXiv.2604.06058}
}

@article{shift2026generative,
  author    = {Rahaman, Kaizer and Deshmukh, Jyotirmoy V. and
               Hota, Ashish R. and Lindemann, Lars},
  title     = {When Environments Shift: Safe Planning with Generative Priors
               and Robust Conformal Prediction},
  journal   = {arXiv preprint arXiv:2602.12616},
  year      = {2026},
  doi       = {10.48550/arXiv.2602.12616}
}

@inproceedings{perez2018film,
  author    = {Perez, Ethan and Strub, Florian and de Vries, Harm and
               Dumoulin, Vincent and Courville, Aaron C.},
  title     = {{FiLM}: Visual Reasoning with a General Conditioning Layer},
  booktitle = {Proceedings of the AAAI Conference on Artificial Intelligence},
  volume    = {32},
  number    = {1},
  year      = {2018},
  doi       = {10.1609/aaai.v32i1.11671}
}

@inproceedings{ha2017hypernet,
  author    = {Ha, David and Dai, Andrew M. and Le, Quoc V.},
  title     = {{HyperNetworks}},
  booktitle = {International Conference on Learning Representations (ICLR)},
  year      = {2017},
  note      = {arXiv:1609.09106},
  doi       = {10.48550/arXiv.1609.09106}
}

@article{korda2018koopman,
  author    = {Korda, Milan and Mezi{\'c}, Igor},
  title     = {Linear predictors for nonlinear dynamical systems:
               {Koopman} operator meets model predictive control},
  journal   = {Automatica},
  volume    = {93},
  pages     = {149--160},
  year      = {2018},
  doi       = {10.1016/j.automatica.2018.03.046}
}

@article{brunton2022koopman,
  author    = {Brunton, Steven L. and Budi{\v{s}}i{\'c}, Marko and
               Kaiser, Eurika and Kutz, J. Nathan},
  title     = {Modern {Koopman} Theory for Dynamical Systems},
  journal   = {SIAM Review},
  volume    = {64},
  number    = {2},
  pages     = {229--340},
  year      = {2022},
  doi       = {10.1137/21M1401243}
}

@article{brunton2016sindyc,
  author    = {Brunton, Steven L. and Proctor, Joshua L. and Kutz, J. Nathan},
  title     = {Sparse Identification of Nonlinear Dynamics with Control
               ({SINDYc})},
  journal   = {IFAC-PapersOnLine},
  volume    = {49},
  number    = {18},
  pages     = {710--715},
  year      = {2016},
  doi       = {10.1016/j.ifacol.2016.10.249}
}

@article{lusch2018koopman,
  author    = {Lusch, Bethany and Kutz, J. Nathan and Brunton, Steven L.},
  title     = {Deep learning for universal linear embeddings of nonlinear
               dynamics},
  journal   = {Nature Communications},
  volume    = {9},
  number    = {1},
  pages     = {4950},
  year      = {2018},
  doi       = {10.1038/s41467-018-07210-0}
}

@article{gu2023mamba,
  author    = {Gu, Albert and Dao, Tri},
  title     = {{Mamba}: Linear-Time Sequence Modeling with Selective State
               Spaces},
  journal   = {arXiv preprint arXiv:2312.00752},
  year      = {2023},
  doi       = {10.48550/arXiv.2312.00752}
}

@inproceedings{smith2023s5,
  author    = {Smith, Jimmy T. H. and Warrington, Andrew and
               Linderman, Scott W.},
  title     = {Simplified State Space Layers for Sequence Modeling},
  booktitle = {International Conference on Learning Representations (ICLR)},
  year      = {2023},
  note      = {arXiv:2208.04933}
}

@article{lindemann2023safe,
  author    = {Lindemann, Lars and Cleaveland, Matthew and Shim, Gihyun and
               Pappas, George J.},
  title     = {Safe Planning in Dynamic Environments Using Conformal
               Prediction},
  journal   = {IEEE Robotics and Automation Letters},
  volume    = {8},
  number    = {8},
  pages     = {5116--5123},
  year      = {2023},
  doi       = {10.1109/LRA.2023.3292071}
}

@inproceedings{yu2020mopo,
  author    = {Yu, Tianhe and Thomas, Garrett and Yu, Lantao and Ermon,
               Stefano and Zou, James and Levine, Sergey and Finn, Chelsea
               and Ma, Tengyu},
  title     = {{MOPO}: Model-based Offline Policy Optimization},
  booktitle = {Advances in Neural Information Processing Systems 33
               ({NeurIPS} 2020)},
  year      = {2020},
  doi       = {10.48550/arXiv.2005.13239}
}

@inproceedings{cleaveland2024robust,
  author    = {Zhao, Yiqi and Hoxha, Bardh and Fainekos, Georgios and
               Deshmukh, Jyotirmoy V. and Lindemann, Lars},
  title     = {Robust Conformal Prediction for {STL} Runtime Verification
               under Distribution Shift},
  booktitle = {Proceedings of the ACM/IEEE 15th International Conference on
               Cyber-Physical Systems (ICCPS)},
  pages     = {169--179},
  year      = {2024},
  doi       = {10.1109/ICCPS61052.2024.00022}
}

@article{falsification2026ocnode,
  author  = {K{\"o}tz, Lasse and Sj{\"o}berg, Jonas and {\AA}kesson, Knut},
  title   = {Optimal Control-Based Falsification of Learnt Dynamics via
             Neural {ODE}s and Symbolic Regression},
  journal = {arXiv preprint arXiv:2602.00031},
  year    = {2026},
  doi     = {10.48550/arXiv.2602.00031}
}

@article{qin2012survey,
  author    = {Qin, S. Joe},
  title     = {Survey on data-driven industrial process monitoring and
               diagnosis},
  journal   = {Annual Reviews in Control},
  volume    = {36},
  number    = {2},
  pages     = {220--234},
  year      = {2012},
  doi       = {10.1016/j.arcontrol.2012.09.004}
}

@article{ching2021softsensors,
  author    = {Ching, Phoebe Mae Lim and So, Richard H. Y. and Morck, Tobias},
  title     = {Advances in soft sensors for wastewater treatment plants:
               A systematic review},
  journal   = {Journal of Water Process Engineering},
  volume    = {44},
  pages     = {102367},
  year      = {2021},
  doi       = {10.1016/j.jwpe.2021.102367}
}

@techreport{iso5469,
  author      = {{ISO/IEC}},
  title       = {{ISO/IEC TR 5469:2024} Artificial intelligence ---
                 Functional safety and {AI} systems},
  institution = {International Organization for Standardization},
  year        = {2024},
  type        = {Technical Report},
  address     = {Geneva, Switzerland},
  url         = {https://www.iso.org/standard/81283.html}
}

\end{document}